\documentclass[10pt,twocolumn,letterpaper]{article}

\usepackage[pagenumbers]{cvpr} % To force page numbers, \eg for an arXiv version

\usepackage[numbers,sort,compress]{natbib}
\usepackage{graphicx}
\usepackage{amsmath}
\usepackage{amssymb}
\usepackage{booktabs}
\usepackage{arydshln}

\usepackage{colortbl}
\usepackage[dvipsnames]{xcolor}
\usepackage[pagebackref,breaklinks,colorlinks]{hyperref}

\usepackage[capitalize]{cleveref}
\crefname{section}{Sec.}{Secs.}
\Crefname{section}{Section}{Sections}
\Crefname{table}{Table}{Tables}
\crefname{table}{Tab.}{Tabs.}

\newcommand{\comment}[1]{}

\newcommand{\R}{\mathbb{R}}

\newcommand{\jac}{\mathbf{D}}

\newcommand{\transval}{\mathbf{y}}
\newcommand{\transvalA}{\transval^{A}}
\newcommand{\transvalB}{\transval^{B}}

\newcommand{\dataval}{\mathbf{x}}
\newcommand{\datavalA}{\dataval^{A}}
\newcommand{\datavalB}{\dataval^{B}}
\newcommand{\datavar}{\mathcal{X}}

\newcommand{\baseval}{\mathbf{z}}
\newcommand{\basevar}{\mathcal{Z}}

\newcommand{\camera}{\mathbf{c}}
\newcommand{\ISO}{\mathbf{g}}

\newcommand{\cleanI}{\mathbf{I}}
\newcommand{\obsI}{\tilde{\cleanI}}
\newcommand{\noiseI}{\mathbf{N}}

\newcommand{\gammaval}{\mathbf{gamma}}

\usepackage{adjustbox}
\usepackage{multicol} 

\def\cvprPaperID{9697} % *** Enter the CVPR Paper ID here
\def\confName{CVPR}
\def\confYear{2022}

\begin{document}

\title{Modeling sRGB Camera Noise with Normalizing Flows}

\author{Shayan Kousha\textsuperscript{1, 3,}\thanks{Work performed while interns at the Samsung AI Center--Toronto.}, Ali Maleky\textsuperscript{1, 3, *}, Michael S. Brown\textsuperscript{3}, Marcus A. Brubaker\textsuperscript{1,2,3}\\
\begin{tabular}{c c c}
    \textsuperscript{1}York University  & \textsuperscript{2}Vector Institute &
    \textsuperscript{3}Samsung AI Center--Toronto
\end{tabular}\\
% {\tt\small [shayanko, \Shayan{ali}, mbrown, mab]@eecs.yorku.ca}
}
\maketitle

%%%%%%%%% ABSTRACT
\begin{abstract}
Noise modeling and reduction are fundamental tasks in low-level computer vision.  They are particularly important for smartphone cameras relying on small sensors that exhibit visually noticeable noise.  There has recently been renewed interest in using data-driven approaches to improve camera noise models via neural networks.  These data-driven approaches target noise present in the raw-sensor image before it has been processed by the camera's image signal processor (ISP).  Modeling noise in the RAW-rgb domain is useful for improving and testing the in-camera denoising algorithm; however, there are situations where the camera's ISP does not apply denoising or additional denoising is desired when the RAW-rgb domain image is no longer available.  In such cases, the sensor noise propagates through the ISP to the final rendered image encoded in standard RGB (sRGB).  The nonlinear steps on the ISP culminate in a significantly more complex noise distribution in the sRGB domain and existing raw-domain noise models are unable to capture the sRGB noise distribution.  We propose a new sRGB-domain noise model based on normalizing flows that is capable of learning the complex noise distribution found in sRGB images under various ISO levels.  Our normalizing flows-based approach outperforms other models by a large margin in noise modeling and synthesis tasks. We also show that image denoisers trained on noisy images synthesized with our noise model outperforms those trained with noise from baselines models. 
\end{abstract}

%%%%%%%%% BODY TEXT

\begin{figure}[ht]
    \centering
    \includegraphics[width=0.99\columnwidth]{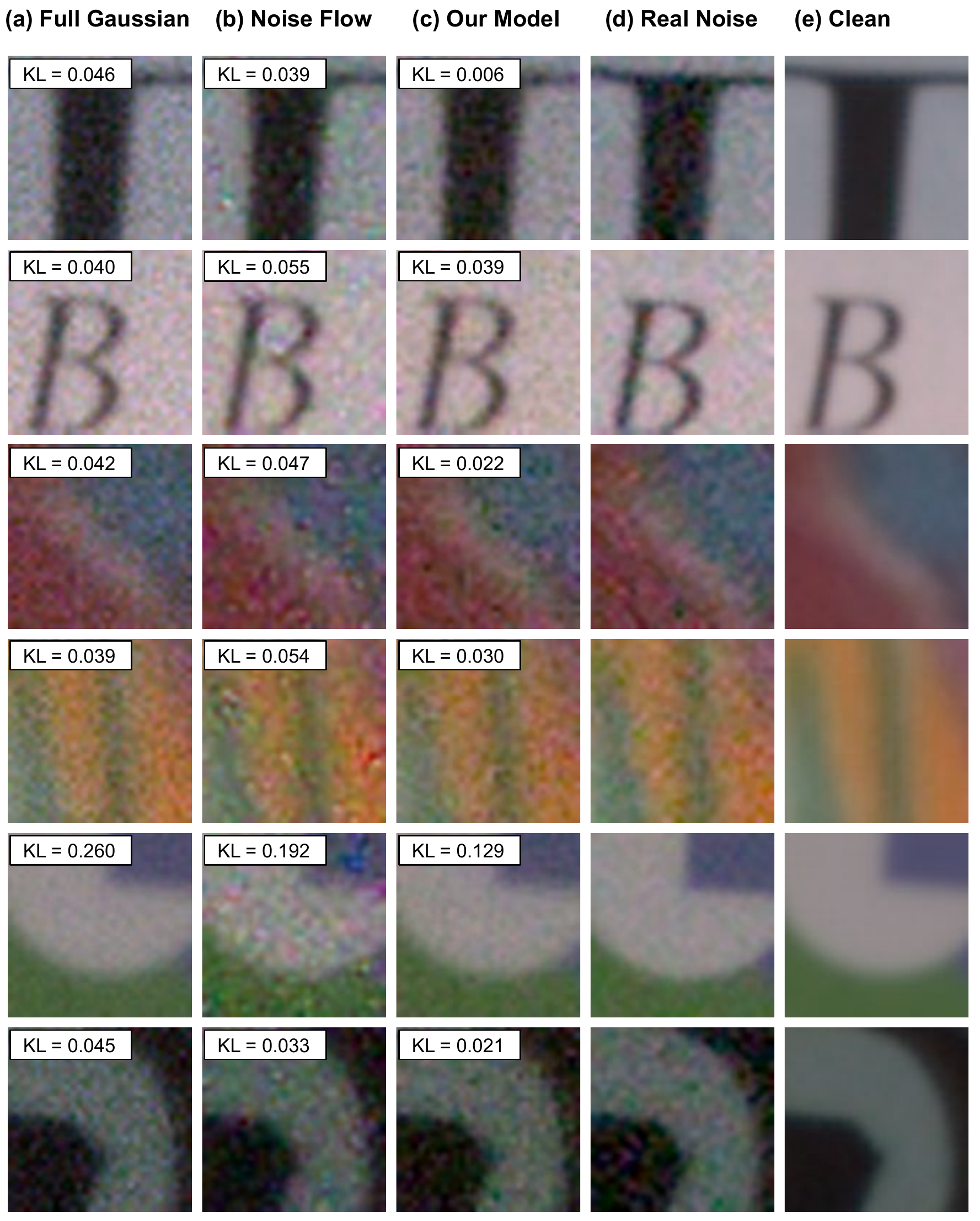}
    \caption{Noisy images generated by (a) a full covariance Gaussian model, (b) Noise Flow \cite{abdelhamed2019noiseflow}, and our model (c) compared with real images (d).  Images are from the SIDD dataset \cite{abdelhamed2018sidd}.
    }
    \label{fig:teaser_fig}
\end{figure}
\begin{figure*}[ht]
    \centering
    \includegraphics[width=0.94\textwidth]{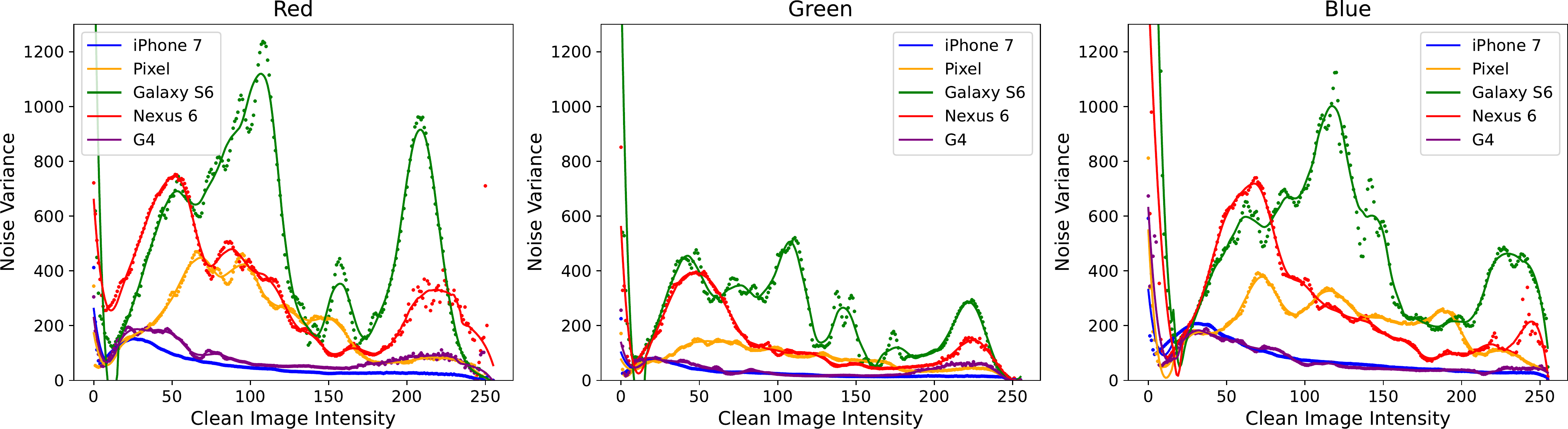}
    \caption{sRGB noise variance as a function of clean image intensity on SIDD \cite{abdelhamed2018sidd}.  These plots exhibit highly variable and unpredictable behavior, in contrast to RAW where there is typically a linear relation between noise variance and intensity. Consequently, we cannot use traditional image noise models like AWGN and NLF and instead propose a novel sRGB focused noise model based on normalizing flows.}
    \label{fig:noise_var}
\end{figure*}

\section{Introduction} \label{sec:intro}

Modeling and reducing noise are long-standing problems in computer vision and image processing with a rich history~(\eg, \cite{Healey1994calibration, Kuan1985adaptive, liu2008automatic}).
While simple models, like simple additive white Gaussian noise (AWGN), have often been used in testing denoising methods, they are well-known not to be realistic and serves only as a rough approximation for real-world camera noise.
When realistic noise models are needed, more sophisticated models, such as Poisson-Gaussian~\cite{Foi2008pg} or Heteroscedastic Gaussian models~\cite{foi2009clipped, liu2014noise}, are used to model the noise distribution observed on camera sensors.
While such models are more realistic than AWGN, they too are often not able to fully capture real camera noise distributions.

In recent years, data-driven noise models have been proposed that learn the noise distribution directly from large datasets of noisy sensor images (\eg, \cite{Nam2016cross_channel, abdelhamed2019noiseflow, chen2018deep, guo2019toward, liu2020gradnet, zhang2017beyond, zhang2018ffdnet}).
These methods focus on modeling the noise present in the raw sensor images.
Modeling noise in the RAW-rgb domain is useful as denoising algorithms are typically applied by the camera's image signal processor (ISP) hardware.  
Such noise reduction is applied early in the ISP's processing pipeline (often in the Bayer processing stages) before the image is rendered to its final sRGB representtion with tonal and color photo-finishing algorithms which are applied to improve the image's perceptual quality.  

It is not unusual, however, for the in-camera denoising to be disabled, not present or insufficient.
Further, most cameras do not save in RAW-rgb by default, making sRGB images far more ubiquitous.
In such cases, the noisy sensor image is rendered through the camera's ISP, including the photo-finishing algorithms that apply complex, nonlinear operations to manipulate the image's tonal and color values.
The resulting noise distribution in the final sRGB is notably more complex than in the unprocessed RAW-rgb space and existing RAW-rgb focused noise models become ineffective for modeling the sRGB noise, as shown in Fig.~\ref{fig:teaser_fig}.  

\paragraph*{Contributions.}
We focus on modeling and synthesizing sRGB image noise, where the in-camera nonlinear processing has altered the noise characteristics from that of the camera's sensor.
We begin with an analysis that shows that existing noise models targeting sensor noise in the RAW-rgb domain are not well suited for sRGB images.
We then propose a generative model that combines recent advances in normalizing flows and captures effects of different gain (ISO) settings and camera types on sRGB image noise.
We show that our sRGB noise model is superior to several baseline noise models.
We further investigate our model's ability to synthesize noise by training a denoiser using noisy images sampled from the model.
We show that this denoiser achieves significantly higher performance compared to denoisers trained on synthesized data from baseline models.

\begin{figure*}[ht]
    \centering
    \includegraphics[width=0.95\textwidth]{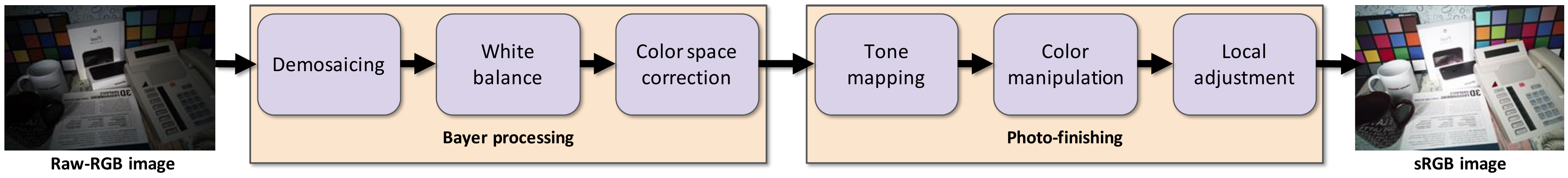}
    \caption{A typical camera ISP processing pipeline. This pipeline processes the RAW-rgb image to encode it in the sRGB domain. The nonlinear steps in the photo-finishing stage substantially complicate the noise distribution in the sRGB domain.}
    \label{fig:incamera_imaging_pipeline}
\end{figure*}

\section{Related work} \label{sec:related work}
Additive white Gaussian noise (AWGN)~\cite{Ohta2001surveillance, rosin1998thresholding, wren1997pfinder} has long been used to model image noise.
% It assumes the noise is from a mean-zero Gaussian distribution with independent and identically distributed values, which is well-known not to reflect real noise observed on camera sensors.
However, it is well known that real camera noise is non-Gaussian, in part because it fails to capture  signal-dependence of the variance.
A common and more realistic model is the heteroscedastic Gaussian model \cite{liu2014noise}, defined as:
\begin{equation} \label{eq:NLF}
    \noiseI \sim \mathcal{N}(0, \beta_1\cleanI + \beta_2),
\end{equation}
where $\beta_1,\beta_2 > 0$ are parameters that model the signal-dependent and signal-independent nature of noise observed on real camera sensors. 
Some cameras include a manufacturer-calibrated heteroscedastic Gaussian noise model in their saved RAW-rgb images, encoded in the DNG format~\cite{abdelhamed2018sidd,abdelhamed2019noiseflow}, although recent work~\cite{zhang2021rethinking} has suggested that such models are often not well calibrated.
The heteroscedastic Gaussian noise model is relatively simple and has few parameters; however, it is still only an approximation of the real sensor noise~\cite{wei2020physicsbased, abdelhamed2018sidd, holst1996ccd, foi2009clipped, plotz2017benchmarking}. 

Researchers have recently begun exploring data-driven approaches.
For example, \citet{abdelhamed2019noiseflow} proposed the Noise Flow model which combined the domain knowledge of signal and gain dependence with the expressiveness of learning-based generative models based on normalizing flows to capture more complex components of the noise.
While the Noise Flow model was effective in simulating RAW-rgb noise, it relied on assumptions that do not apply in the sRGB color space.
For example, the Noise Flow model builds off the heteroscedastic Gaussian model, which assumes the noise variance linearly depends on the underlying clean image intensity.
However, this assumption no longer applies in the sRGB image domain (see Fig.~\ref{fig:noise_var}) due to subsequent non-linear and potentially content-dependent processing of the RAW-rgb image.
As a result, in our experiments we show that simply applying the Noise Flow model to sRGB data fails to capture the noise distribution.
% This noise model contains fewer than 2500 parameters and does not have the power to capture complex noise in the sRGB domain.
% We have modified the code of Noise Flow to operate in the sRGB domain without affecting the behavior of any of the transformations.
% We use it as one of our baselines to show models designed for RAW data cannot directly be applied to sRGB images because these models do not consider non-linearities introduced by in-camera image processing steps such as Gamut mapping and Tone mapping.

There have been relatively few attempts to model noise from sRGB data.
\citet{Nam2016cross_channel} introduced a model that is designed for the specific use case of modeling noise and other degradations caused by JPEG compression.
More recently, the C2N \cite{Jang2021C2n} model attempted to model noise using unpaired clean and noisy images using a generative adversarial network (GAN).
% Unfortunately due to the nature of GANs, evaluating its noise generation capabilities directly is challenging.
% However, we use this model as one of our baselines.

In this paper, we introduce a data-driven model for image noise in the sRGB domain that conditions on camera settings and the underlying clean image.
Like Noise Flow \cite{abdelhamed2019noiseflow}, the model is built using normalizing flows \cite{dinh2017density, kingma2018glow, kobyzev2021review} but avoids making unrealistic assumptions that apply only to RAW-rgb.
The resulting model is shown to have state-of-the-art noise modeling capabilities for sRGB noise.
Further, when training a denoiser using noisy samples from the proposed noise model, we show that the resulting denoiser significantly out-performs those trained with existing sRGB noise models.

\section{Preliminaries}
\label{sec:background}

Image noise is a combination of degradations introduced by multiple sources in the imaging process and physical limitations of sources like camera sensors.
Many of these sources occur when first capturing the RAW-rgb image by the camera sensors.
Subsequently, the RAW-rgb image (including noise) is subject to in-camera imaging process that transforms the image from the scene-referred RAW-rgb color space to the display-referred sRGB color space.
Fig.~\ref{fig:incamera_imaging_pipeline} summarizes some of the main steps in the in-camera imaging pipeline.
The results of the photo-finishing steps, many nonlinear in nature, introduce new sources of noise (\eg, from clipping and sharpening) while amplifying and distorting the original sensor's noise distribution.
The noise from these multiple sources can be characterized as:
\begin{equation}
    \label{eq:noise_eq}
    \obsI = \cleanI + \noiseI,
\end{equation}
where $\obsI$ is the observed, noisy image, $\cleanI$ is the true, underlying clean image, and $\noiseI$ is the the noise whose distribution may depend on $\cleanI$.
In this work, we aim to design a generative model that captures the complexity of noise introduced by all sources.

\subsection*{Normalizing Flows}
Normalizing flows are a family of generative models that have gained popularity in recent years.
Due to their formulation, they admit both efficient sampling and exact evaluation of probability density in contrast to other generative models, like GANs and VAEs \cite{bond2021deep, kobyzev2021review}.
Additionally, flow-based models do not suffer from issues like mode and posterior collapse that are commonly faced when training GANs and VAEs.

Below we briefly introduce normalizing flows.
We refer the reader to recent review articles \cite{kobyzev2021review,papamakarios2021normalizing} for a more extensive treatment.
A normalizing flow consists of differentiable and bijective functions that learn a transformation $\baseval = f(\dataval|\Theta)$ with parameters $\Theta$ called a \emph{flow}.
% $\baseval, \dataval \in \R^{d}$,
A flow transforms data samples $\dataval \in \R^d$ from a complex distribution, $p_\datavar$, to some base space $\baseval \in \R^d$ with a known and tractable distribution and probability density function, $p_\basevar$.
Here, as is common, we will assume that $p_\basevar$ takes the form of an isotropic Gaussian distribution with unit variance.
The probability density function in the data space can then be found using the change of variables formula
\begin{equation}
    p_\datavar(\dataval) = p_\basevar(f(\dataval|\Theta))\left|\det \jac f(\dataval|\Theta) \right|,
\end{equation}
where $\jac f(\dataval)$ is the Jacobian matrix of $f$ at $\dataval$.
The result is a model that, given a dataset $D = \{\dataval_{i}\}_{i=1}^{M}$, can be trained by using stochastic gradient descent to minimize the negative log likelihood of the data
\begin{equation}
-\sum_{i=1}^M \log p_\basevar(f(\dataval_i|\Theta))+\log \left|\det \jac f(\dataval_i|\Theta) \right|,
\label{eq:log_likelihood}
\end{equation}
with respect to the parameters $\Theta$.
Samples can be generated by sampling from the base distribution $\baseval \sim p_\basevar$ and then applying the inverse flow $\dataval = f^{-1}(\baseval)$.

Formally, normalizing flows define distributions over continuous spaces.
To apply them to quantized data (\eg, as sRGB data which is typically truncated to 256 intensity levels) some attention must be paid to avoid a degeneracy \cite{theis2016note} that occurs when fitting continuous density models to discrete data.
Here we use uniform dequantization, which adds uniformly sampled noise to the images during training.
More complex forms of dequantization are possible \cite{ho2019flow}.

Constructing expressive, differentiable, and bijective functions is the primary research problem in normalizing flows, and there have been many attempts at this; see \cite{kobyzev2021review,papamakarios2021normalizing} for a thorough review.
A flow $f$ is typically constructed by the composition of simpler flows---namely, $f = f_{1} \circ ... \circ f_{N - 1} \circ f_{N}$---since the composition of bijective functions is itself bijective.
Analogous to adding depth in a neural network, composing flows can increase the complexity of the resulting distribution $p_\datavar$.
Individual flows are typically constructed such that their inverse and Jacobian determinant are easily calculated.
Next we review two common forms of bijection that we will use.
% There have been many works trying to define transformations that are easily invertible and their determinant of the Jacobian is trivially obtained \cite{kingma2018glow, dinh2015nice, ho2019flow, dinh2017density}. We use the following transformations introduced in these papers as our starting point:

\paragraph*{Affine Coupling}
% \noindent{\textbf{Affine Coupling}~\cite{dinh2017density}}:
Affine coupling flows \cite{dinh2017density} are a simple, efficient and widely used form of flow.
They work by splitting the input dimensions, $\dataval = (\datavalA,\datavalB)$, into two disjoint subsets, $\datavalA,\datavalB$.
Then, one subset, $\datavalA$, is unmodified but used to compute scale and translation factors, which are applied to the other subset, $\datavalB$.
Formally, an affine coupling layer is defined as $\transval = (\transvalA,\transvalB)$, where
$\transvalA = \datavalA$ and
\begin{equation*}
\transvalB = \datavalB \odot f_s(\datavalA|\Theta) + f_t(\datavalA|\Theta),
\end{equation*}
where $\odot$ is the element-wise product.
The functions $f_s$ and $f_t$ compute the scale and translation factors and can be arbitrary, for example, deep neural networks.
The inverse of this layer is easily computed as $\datavalB = \left(\transvalB - f_{t}(\transvalA|\Theta)\right) \oslash f_{s}(\transvalA|\Theta)$.
Further, the log determinant of this transformation is efficiently calculated as $\sum \log f_s(\datavalA|\Theta)$, where the sum is taken over the output dimensions of $f_s$.

\paragraph*{1x1 Convolution}
% \noindent{\textbf{Invertible 1x1 Convolution}~\cite{kingma2018glow}}:
Coupling layers must change the way dimensions are split between layers.
This can be done by a random permutations \cite{dinh2015nice,dinh2017density} but the Glow model \cite{kingma2018glow} introduced the use of 1x1 convolutions as an invertible transformation.
In essence, these layers are full linear transformations applied channel-wise to the inputs.
The inverse is simply the inverse linear transformation applied channel-wise and the log determinant term is the number of pixels times the log determinant of the linear transformation.
%We found the LU decomposed parameterization of this transformation to help with the stability of our model.

\begin{figure}[ht]
    \centering
    \includegraphics[width=0.8\columnwidth]{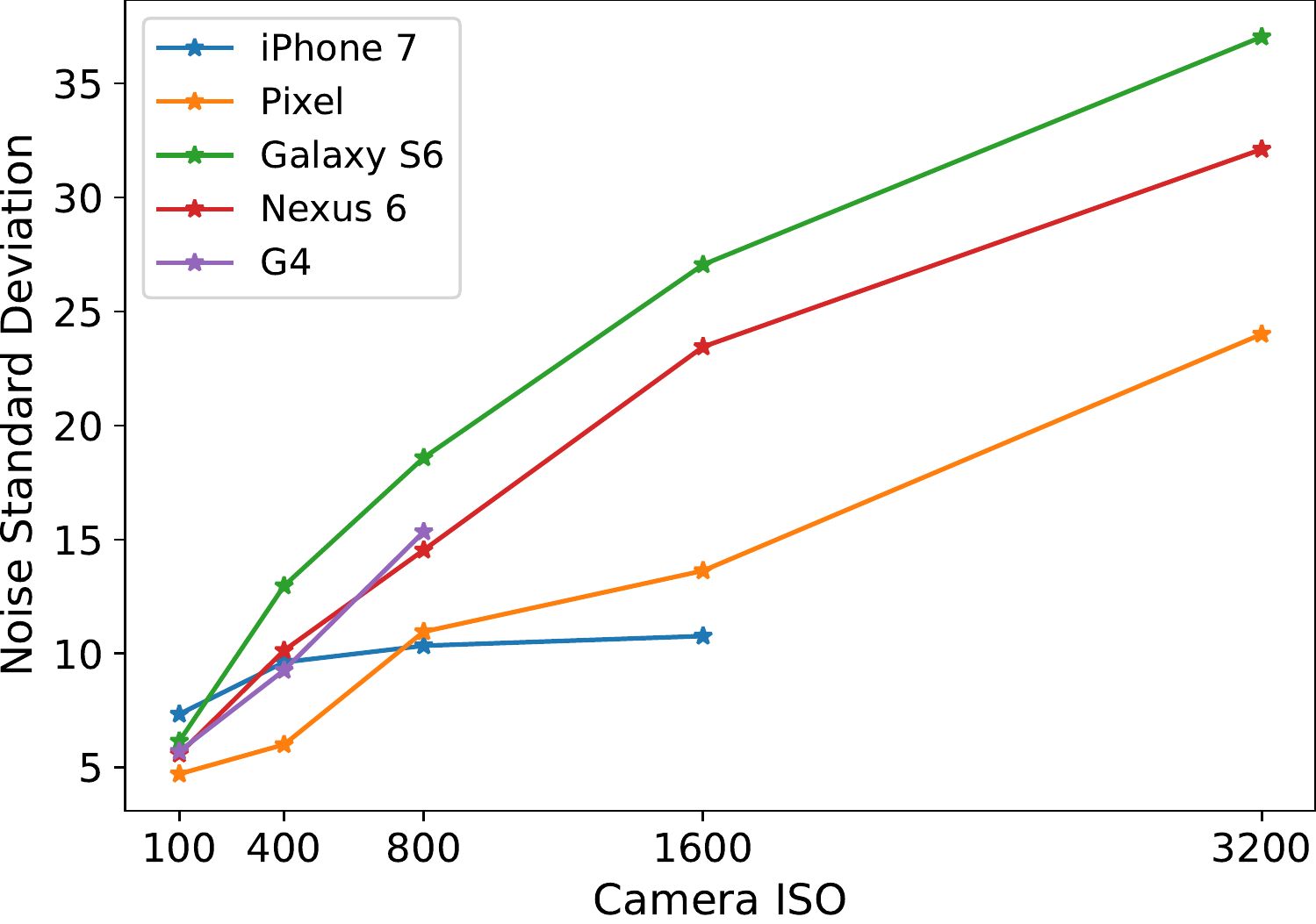}
    \caption{Real noise standard deviation changes as the sensitivity of the camera's sensor increases. The camera type is another factor affecting the noise behavior. For example, the noise standard deviation of images captured by Galaxy S6 is the highest under almost all ISO levels. Analysis done on the SIDD \cite{abdelhamed2018sidd}.}
    \label{fig:cam_std}
\end{figure}
\begin{figure*}[ht]
    \centering
    \includegraphics[width=0.99\textwidth]{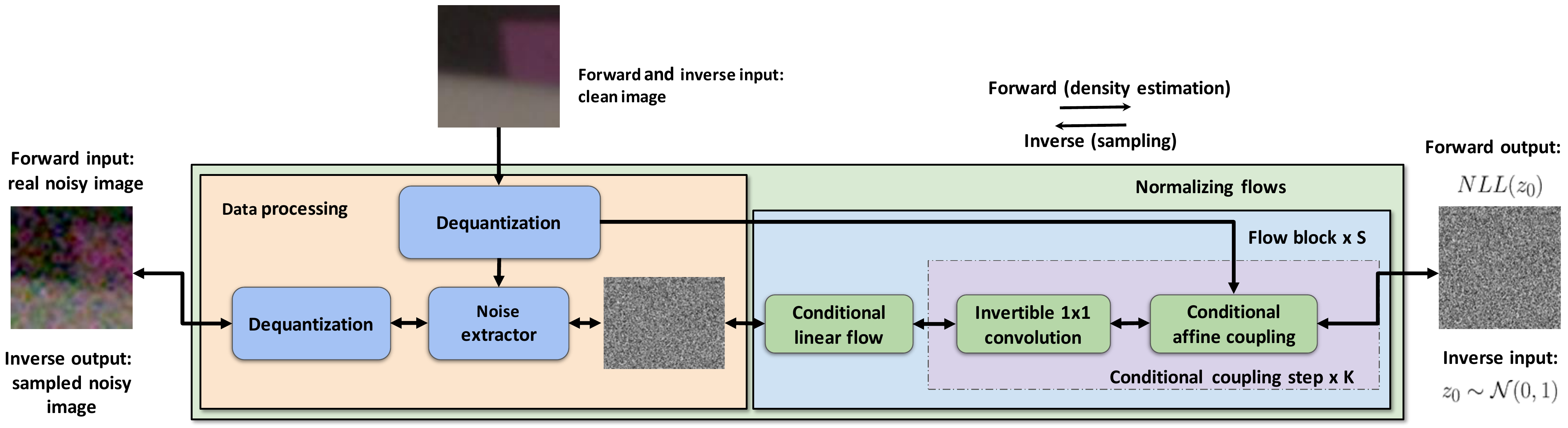}
    \caption{Our model consists of two main sections: (1) The data processing section is responsible for processing noisy and clean images. (2) The flow steps are responsible for learning the complex noise distribution.}
    \label{fig:model_architecture}
\end{figure*}

\section{Normalizing Flows for sRGB Noise} \label{sec:Method}

Here we introduce our model of sRGB noise based on normalizing flows.
% The model is designed based on our understanding of noise behavior under different conditions and settings.
Our analysis on the SIDD dataset \cite{abdelhamed2018sidd}, summarized in Fig.~\ref{fig:noise_var}, confirms that real noise in the sRGB domain has a complex structure that is not captured by the standard, heteroscedastic-based noise models and varies significantly between cameras and even color channels.
Inspired by this analysis, we introduce two new conditional flows in the following sections that allow the noise model to be conditioned on critical parameters, such as camera model, gain setting and clean intensity.

Figure \ref{fig:model_architecture} shows the proposed model architecture.
Here we describe it as a transformation from the data (\ie, a noisy, observed sRGB image $\obsI$) to the base space $\baseval$.
First, input images are dequantized with uniform dequantization as mentioned above.
Unlike RAW-rgb, which is typically represented as a floating point number, sRGB data is typically quantized to 256 intensity levels.
Note that, because both the clean and observed images have been quantized, both need to be dequantized.
Next, the clean image, $\cleanI$, is subtracted from the observed image, $\obsI$, to get the noise image, $\noiseI$.
This is then followed by $S$ flow blocks, which are responsible for learning a transformation from the noise to a sample in the base distribution, and vice versa.
These flow blocks consist of one conditional linear (CL) flow followed by $K$ conditional coupling steps (CCS), where a conditional coupling step consists of one invertible 1x1 convolution layer and one conditional affine coupling transformation.
In our experiments, we use $S = 4$ and $K = 2$, unless otherwise specified.
Next, we describe the conditional linear flow and conditional affine coupling layers.

\subsection{Conditional Linear Flow}
The nature of the noise and the subsequent non-linear processing is heavily determined by the specific camera and gain (or ISO) settings used.
To account for this, we introduce a linear flow layer, which is conditioned on the camera, $\camera$, and gain setting, $\ISO$, of the camera.
This has the form
\begin{equation}
    \transval = \dataval \odot f_{s}(\camera, \ISO) + f_{t}(\camera, \ISO),
\end{equation}
where $\odot$ is the element-wise product, $\dataval$ is the input, $\transval$ is the output, and $f_{s}$ and $f_{t}$ are functions that output the scale and translation factors.
The functions $f_{s}$ and $f_{t}$ can be arbitrarily complex and have no constraints other than that $f_{s} \neq 0$.
See supplemental materials for architectural details of $f_s$ and $f_t$.
% In practice, they are, themselves, a linear transformation of a one-hot encoding of $\camera$ and $\ISO$ which produces a scale and translation output for each channel, followed by an exponential function applied to the scale to ensure the output of $f_{s}$ is positive.
The inverse of this layer is easily calculated as:
\begin{equation}
    \dataval = \left(\transval - f_{t}(\camera, \ISO)\right) \oslash f_{s}(\camera, \ISO),
\end{equation}
where $\oslash$ is element-wise division.
The log-determinant is given by $\sum \log f_{s}(\camera, \ISO)$, where the sum is taken over all dimensions of the input.

\subsection{Conditional Affine Coupling}
This layer is an extension of the affine coupling layer presented above.
To capture the complex dependence of the noise distribution on both the underlying clean image and the camera and gain settings (see Figs.~\ref{fig:noise_var} and \ref{fig:cam_std}), we extend the coupling layers to take these values as input.
The conditional affine coupling layer is similar to the standard affine coupling layer but differs in the the scale and translation factors.
Specifically, the output of the layer is $\transval=(\transvalA,\transvalB)$ with $\transvalA = \datavalA$ and
\begin{equation*}
\transvalB = \datavalB \odot f_{cs}(\datavalA|\cleanI,\camera,\ISO) 
    + f_{ct}(\datavalA|\cleanI,\camera,\ISO),
\end{equation*}
where $\odot$ is the element-wise product, $\dataval = (\datavalA, \datavalB)$ is the input, and $f_{cs}$ and $f_{ct}$ are functions that compute the conditional scale and translation based on the input clean image, $\cleanI$, camera, $\camera$, and gain setting, $\ISO$.
The inverse and log determinant of this transformation can be easily calculated, analogous to the (unconditional) coupling layer.
See the supplemental material for architectural details of $f_{cs}$ and $f_{ct}$.

\section{Experiments}\label{sec:experiments}
To evaluate our model we use the SIDD dataset \cite{abdelhamed2018sidd}.
The SIDD-Medium split contains 320 noisy-clean image pairs captured under various ISO and lighting conditions taken by five different smartphones.
While this dataset provides the data in both RAW and sRGB domains, here, we use only sRGB images.
Note that this dataset used a simplified software ISP to render images from the captured RAW-rgb to sRGB instead of directly using the sRGB images produced by the camera. The software ISP applies the camera parameters stored in the raw’s DNG file (\eg, white-balancing, lens shading correction, color space mapping, custom tone-map, mapping to sRGB, and sRGB gamma).
We extract approximately 3,000 patches of size 32x32 from each image.
From these extracted patches, 80\% are used for training and the remaining for validation.
Patches are randomly distributed to ensure all cameras and ISO settings are fairly represented in both training and validation sets.
For training we minimize the negative log likelihood (Eq.~\ref{eq:log_likelihood}) using the Adam optimizer \cite{kingma2015adam}.

\subsection{Metrics}
To quantitatively evaluate the model we consider two metrics.
First, the negative log likelihood per dimension (NLL) on the test set is used as a direct evaluation of density estimation.
Second, to better assess the quality of the sampled noise, we use the  Kullback-Leibler (KL) divergence which was introduced in \cite{abdelhamed2019noiseflow}.
This metric computes the KL divergence between histograms of real and sampled noise.
This metric is more sensitive to mismatches in the model's estimated variance than the NLL metric is.
% Considering that the characteristic of noise data varies a lot based on the camera and ISO level used to capture the image, KL divergence is a good metric to evaluate our model’s ability to capture the basic characteristics of the distribution.
% Compared to NLL, KL divergence is more sensitive to the mean and variance of the generated sample.

\subsection{Baselines}
We explored a number of baseline sRGB noise models.
We considered three variations of homoscedastic Gaussian noise: 1) AWGN, which assumes independent, isotropic noise at each pixel; 2) diagonal covariance Gaussian, which assumes independent but anisotropic noise at each pixel; and 3) full covariance Gaussian, which allows correlations between color channels.
Note the full covariance Gaussian model was previously proposed for sRGB data by \cite{Nam2016cross_channel}.
We also implemented a heteroscedastic Gaussian model, often referred to as the noise level function (NLF) and described in Eq.~\ref{eq:NLF}.
While not expected to perform well based on, \eg, Figure \ref{fig:noise_var}, it is a widely used and well known model of camera noise.
Finally, we compared with a direct adaption of the Noise Flow model \cite{abdelhamed2019noiseflow} to sRGB instead of RAW-rgb data.
To do this we modified the number of channels that the architecture expected, but otherwise left it unchanged.
Because of the strong assumptions made, Noise Flow has a small number of parameters. To make a more fair comparison, we also built a larger version of Noise Flow, referred to as Noise Flow-Large, that follows the architecture of Noise Flow but with the number of parameters increased to be comparable to our model.
All baselines were implemented as normalizing flows using combinations of simple linear flows, signal-dependent flows, and gain-dependent flows  \cite{abdelhamed2019noiseflow} to ensure consistency.

\begin{figure}[t]
    \centering
    \subfloat[\centering]
    {{\includegraphics[width=0.475\columnwidth]{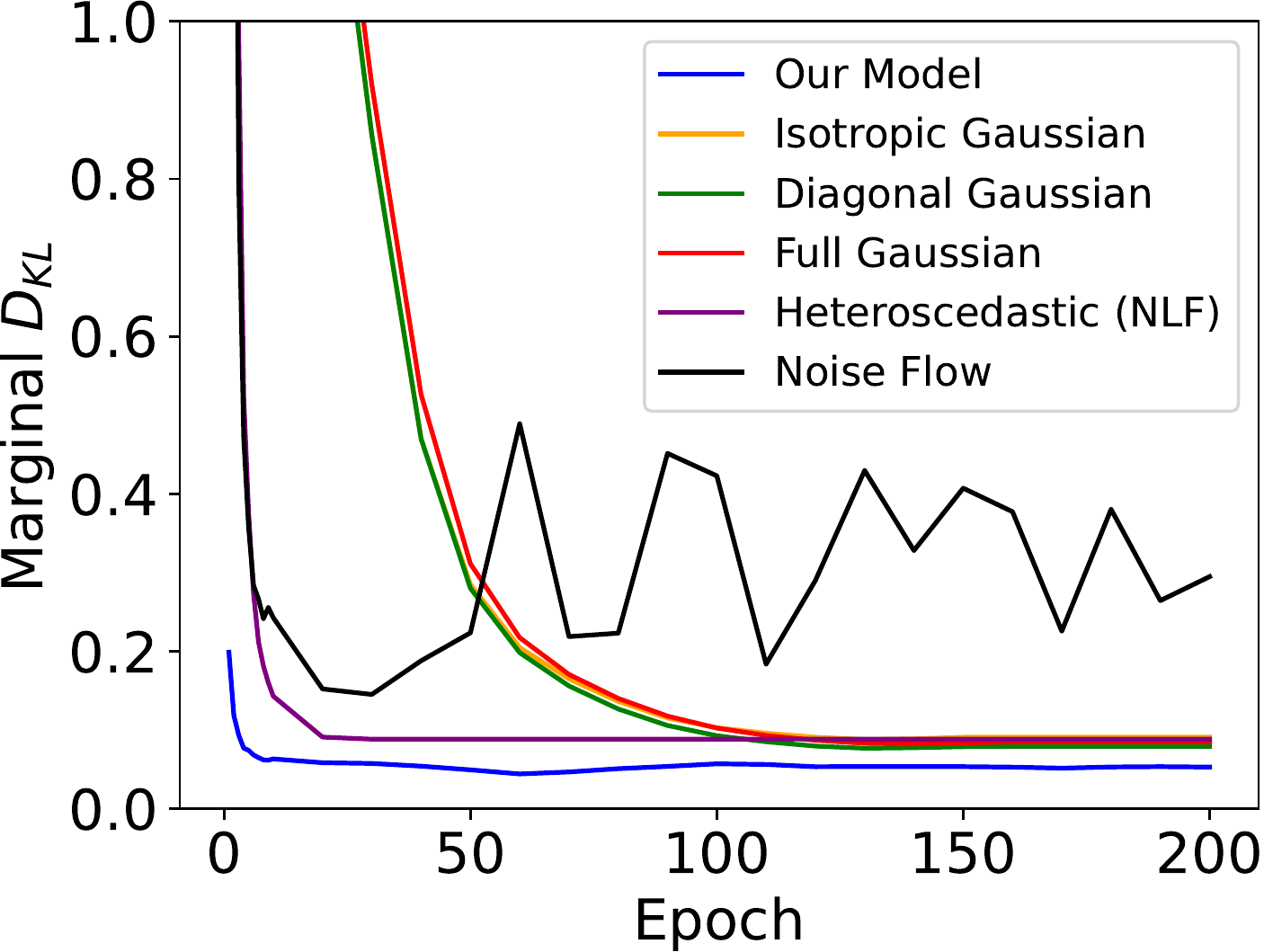} }\label{fig:test_kld}}
    \qquad
    \hskip -4ex
    \subfloat[\centering ]
    {{\includegraphics[width=0.475\columnwidth]{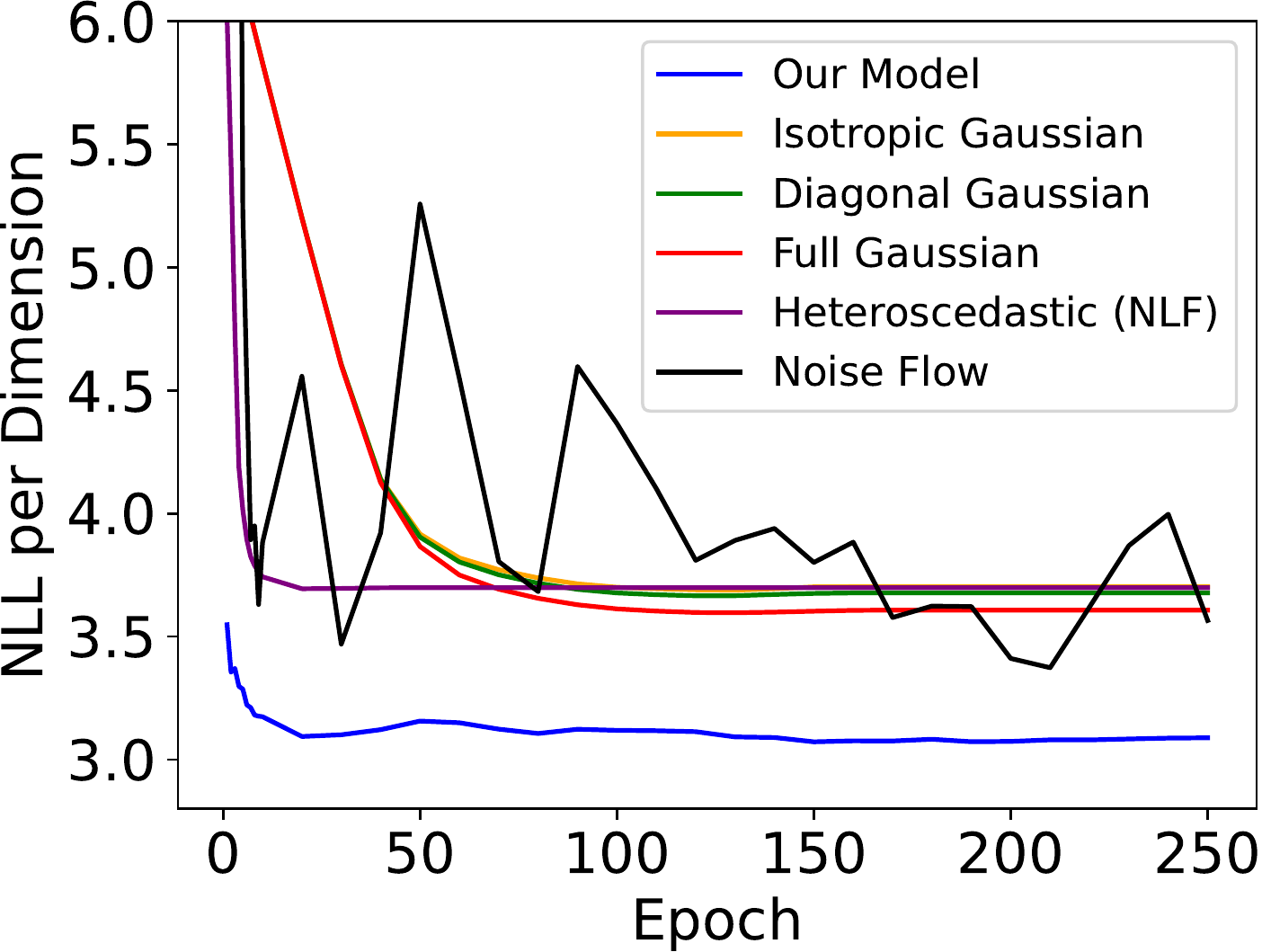} } \label{fig:test_nll}}
    \caption{(a) Marginal KL divergence between the synthetic noise from the models and the real noise samples of the test set. (b) Testing NLL per dimension of our model compared to our baseline models.
    Our model not only performs by both metrics, it also converges faster.}
    \label{fig:kld_nll}
\end{figure}

\begin{table}[t]
\centering
% \begin{adjustbox}{width=\columnwidth,center}
\begin{tabular}{ c c c c }
    \hline
    Model & $NLL$ & $D_{KL}$ & $\# Params$ \\
    \hline
     Isotropic Gaussian & 3.703 & 0.091 & 50 \\  
     Diagonal Gaussian & 3.678 & 0.079 & 150\\ 
     Full Gaussian & 3.608 & 0.085 & 525 \\
     Heteroscedastic (NLF)  & 3.642 & 0.088 & 72\\
     Noise flow~\cite{abdelhamed2019noiseflow} & 3.311 & 0.198 & 2330\\
     Noise Flow-Large & 3.288 & 0.227 & 6618 \\
     \hline
     Our model & \textbf{3.072} & \textbf{0.044} & 6160 \\
     \hline
\end{tabular}
% \end{adjustbox}
\caption{Test $NLL$ and marginal $D_{KL}$ for our model and our baselines. The proposed model outperforms the baselines in both metrics by a large margin. Noise Flow-Large and Noise Flow models have the closest NLL to the proposed method, however, their relatively high $D_{KL}$ indicates that these models fail to generate realistic noise samples.
% \Shayan{The larger Noise Flow model follows the architecture of Noise Flow but has the same number of parameters as our proposed architecture, we see that while it marginally improves performance over the baseline Noise Flow model, it is still significantly outperformed by our proposed architecture.}
Together the results show that models originally developed for RAW-rgb are unlikely to be successful with sRGB. }
\label{tab:nll_kld}
\end{table}

\subsection{Results}

Table \ref{tab:nll_kld} shows the final test NLL and KL divergence for our model and all the baselines.
Figures \ref{fig:test_kld} and \ref{fig:test_nll} show the KL divergence and NLL during training of all models.
The results demonstrate that our model achieves a lower (better) NLL (3.072 vs. 3.311 nats/pixel for Noise Flow), and converges faster, requiring just a few epochs of training.
During training the NLL of the Noise Flow model fluctuates significantly compared to the other baselines.
We believe this is due to the assumptions made in the model architecture, specifically the signal-dependent layer, which do not hold in the sRGB domain and further demonstrate the need for a different approach to noise modeling in the sRGB domain.

In terms of marginal KL divergence our model also significantly improves over the baselines.
Unlike with NLL, the closest performing baseline in terms of KL divergence was the diagonal Gaussian noise model, with a KL divergence of $0.079$.
In contrast, our model achieved a KL divergence of $0.044$.
For comparison, we also considered the recently proposed C2N \cite{Jang2021C2n} model, a GAN-based sRGB noise model.
Their paper reported a KL divergence of $0.1638$. However, we note that KL divergence is sensitive to the choice of histogram bins and other implementation details.
% At the time of submission, the C2N code was unavailable and a direct comparison was not possible.

\begin{figure}[t]
    \centering
    \includegraphics[width=0.8\columnwidth]{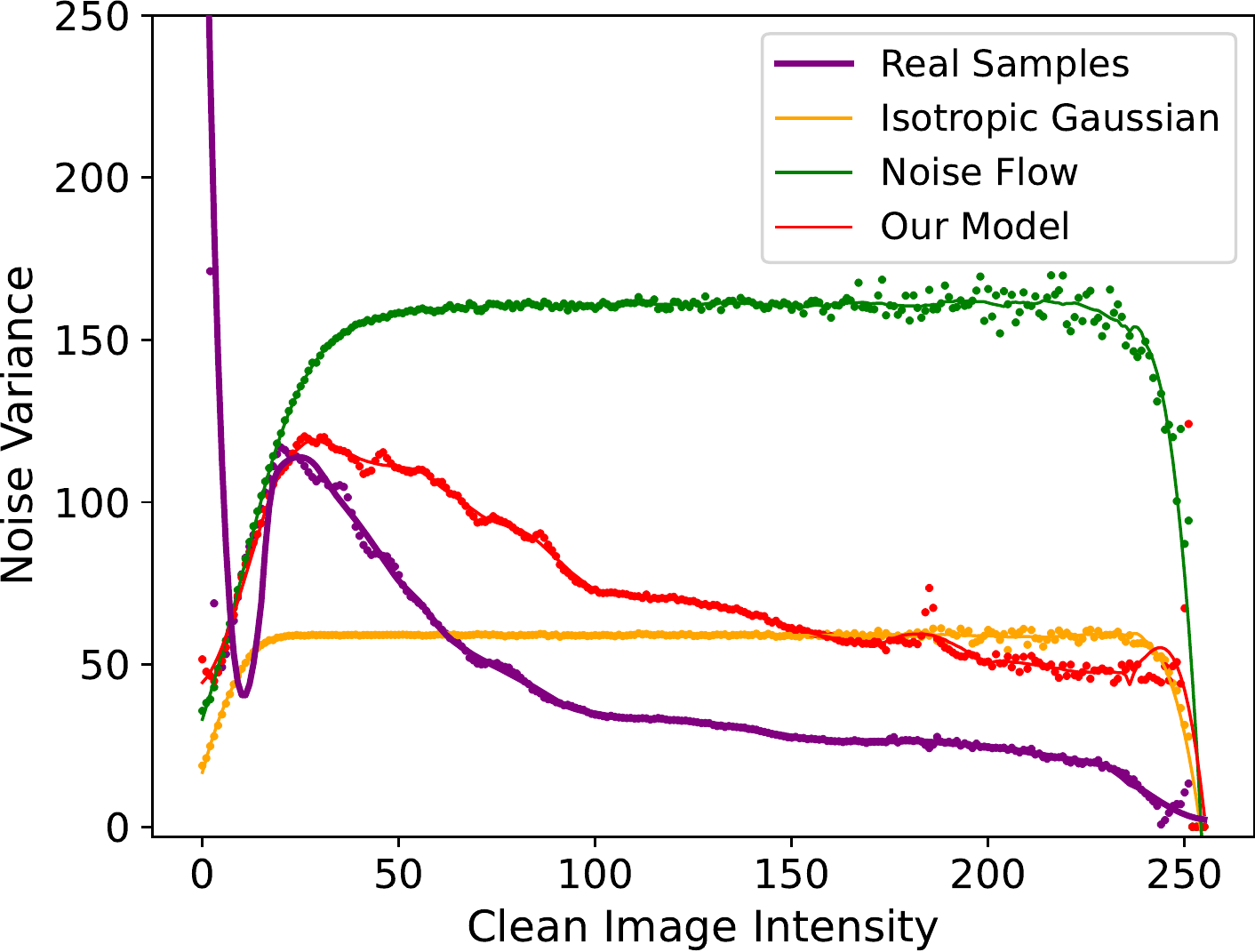}
    \caption{Red channel noise variance of real noise samples from iPhone7-ISO 100 setting and samples generated from our model and two of the baselines given clean images corresponding to the real noise samples. Our model success fully shows a similar noise variance trend to the real noise. However, the Noise Flow and Isotropic Gaussian models fail to learn this trend.}
    \label{fig:red_var}
\end{figure}
\begin{figure}[t]
    \centering
    \includegraphics[width=\columnwidth]{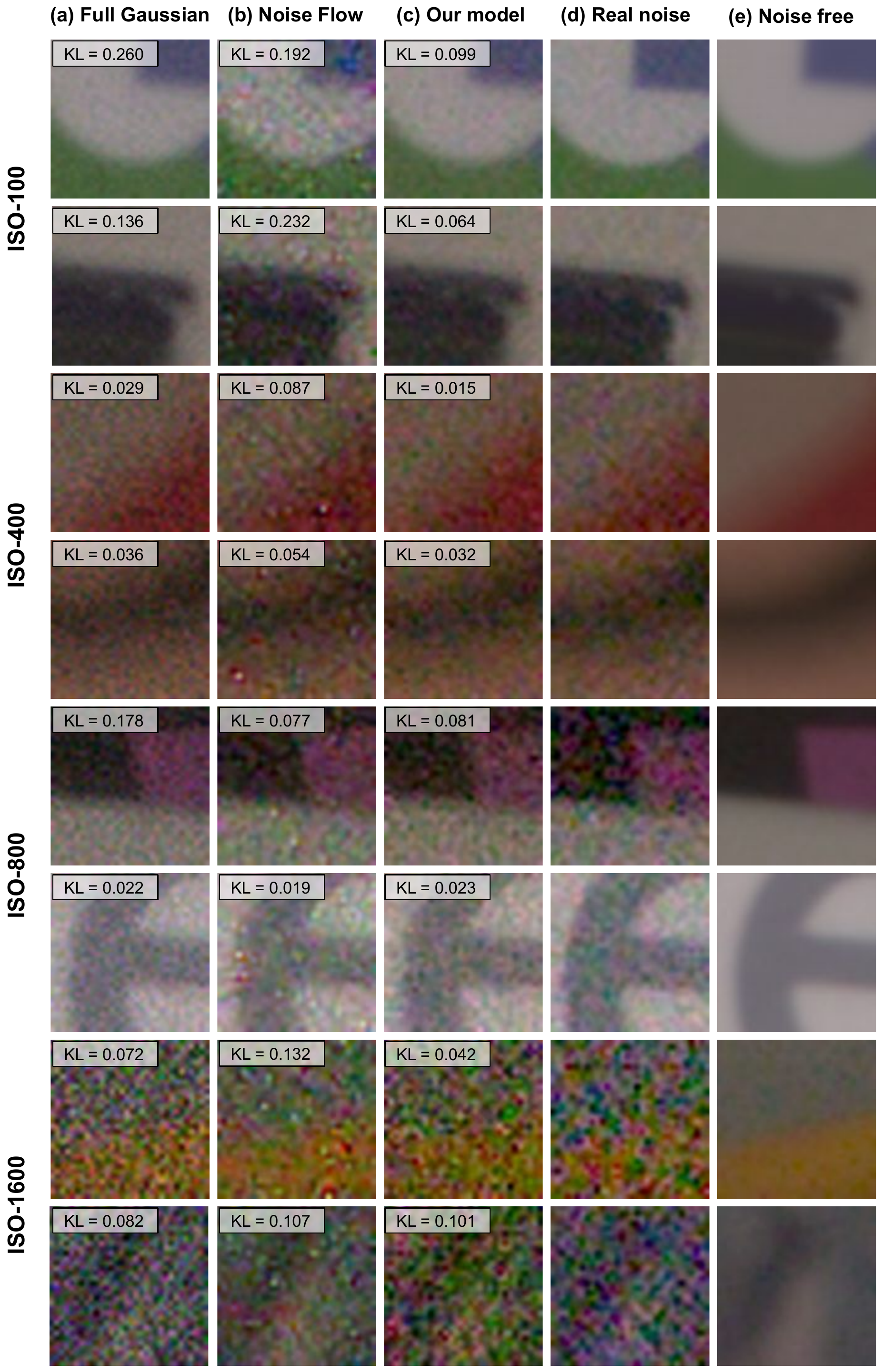}
    \caption{Generated samples from our model and two baselines. Samples from our model have noticeable visual similarities with the real noisy samples. Our samples achieve the lowest $D_{KL}$ in almost all cases, showing its ability to generate realistic noise.}
    \label{fig:noisy_images}
\end{figure}

Interestingly, Table \ref{tab:nll_kld} shows that Noise Flow significantly outperformed the other baseline noise models in terms of NLL, but significantly underperformed them in terms of KL divergence.
On further investigation, we found that the Noise Flow model was significantly overestimating the variance of the noise in many cases.
For instance, Figure \ref{fig:red_var}, similar to Figure \ref{fig:noise_var}, shows the variance of sRGB noise as a function of noise-free image intensity for real data, our model, Noise Flow, and the isotropic baseline for an iPhone7 with ISO level of 100.
This graph shows that Noise Flow has badly overestimated the variance, even compared to the isotropic Gaussian model, suggesting that the difficulties in training seen earlier are preventing it from converging to a reasonable model.
In contrast, while our proposed model slightly overestimates the noise as well, it much better captures the structure of the relationship.
% samples generated from our model tend to follow the characteristics of the real noise samples. Our baseline models, namely Additive White Gaussian Noise and Noise Flow models, are mainly flat and their generated noise variance does not change significantly as the clean image intensity increases.
% The line of the AWGN model in the figure is not fully flat as expected due to the effects of clipping done to make sure the values of noisy samples are between 0 and 255.

\paragraph{Qualitative Comparison.}
To qualitatively compare the trained noise models Figure \ref{fig:noisy_images} shows samples of noise from our model and with two baseline models at different ISO levels.
Out of the Gaussian-based models, the full covariance Gaussian Model achieves the best test NLL and has been shown to be a good fit for modeling noise in the sRGB space as its full covariance can learn dependencies between channels~\cite{Nam2016cross_channel}.
A more extensive set of samples is available in the supplemental materials.
Samples from our model are generally more visually similar to those of real noisy images, particularly in comparison to the baselines.
Noise Flow samples are too noisy and samples from full covariance Gaussian do not exhibit enough variance.
For example, Noise Flow samples at ISO 800 are significantly less noisy compared to the real noisy image.

\begin{figure}[t]
    \centering
    \subfloat[\centering]
    {{\includegraphics[width=0.48\columnwidth]{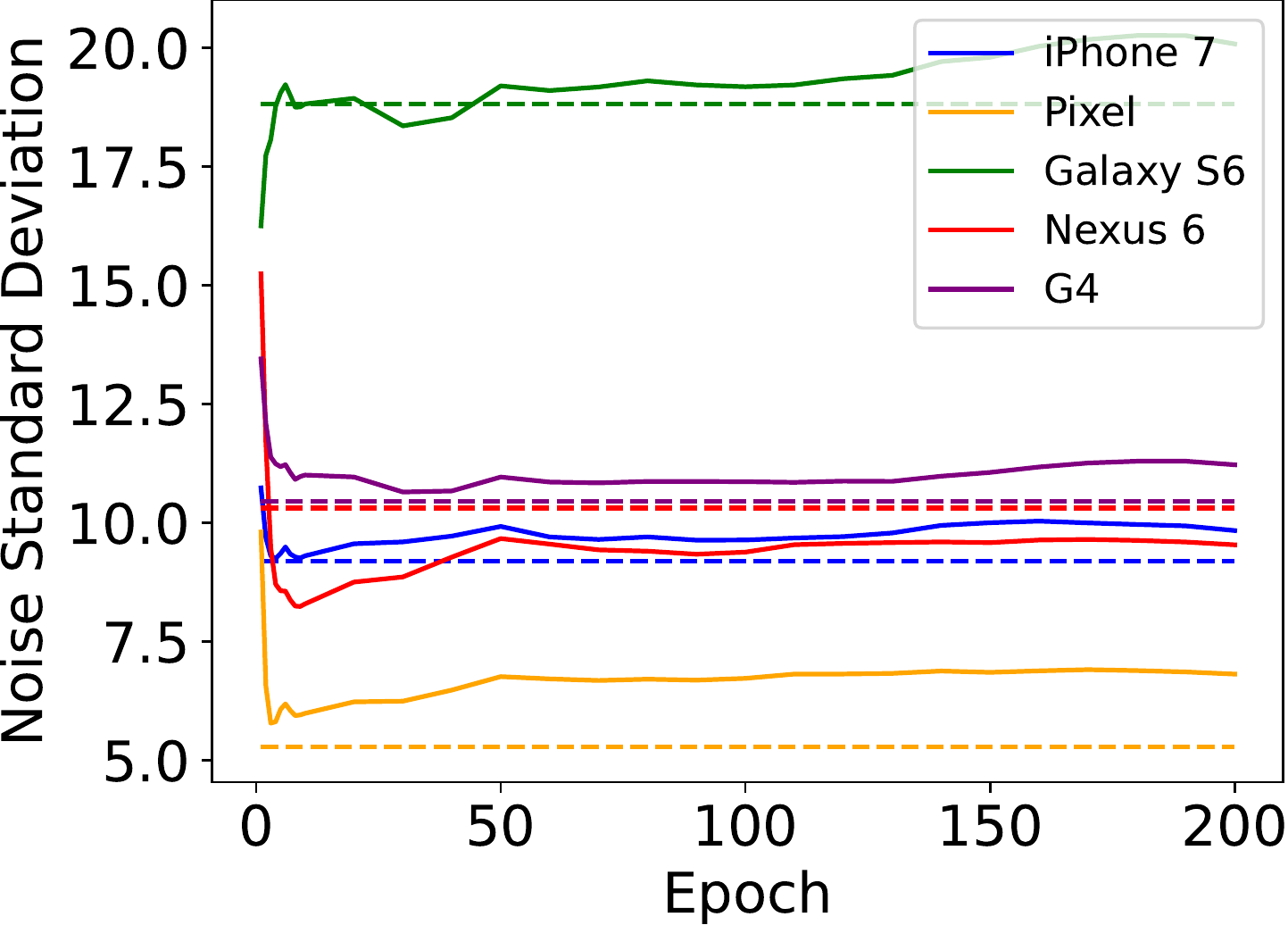} }}
    \qquad
    \hskip -4ex
    \subfloat[\centering]
    {{\includegraphics[width=0.48\columnwidth]{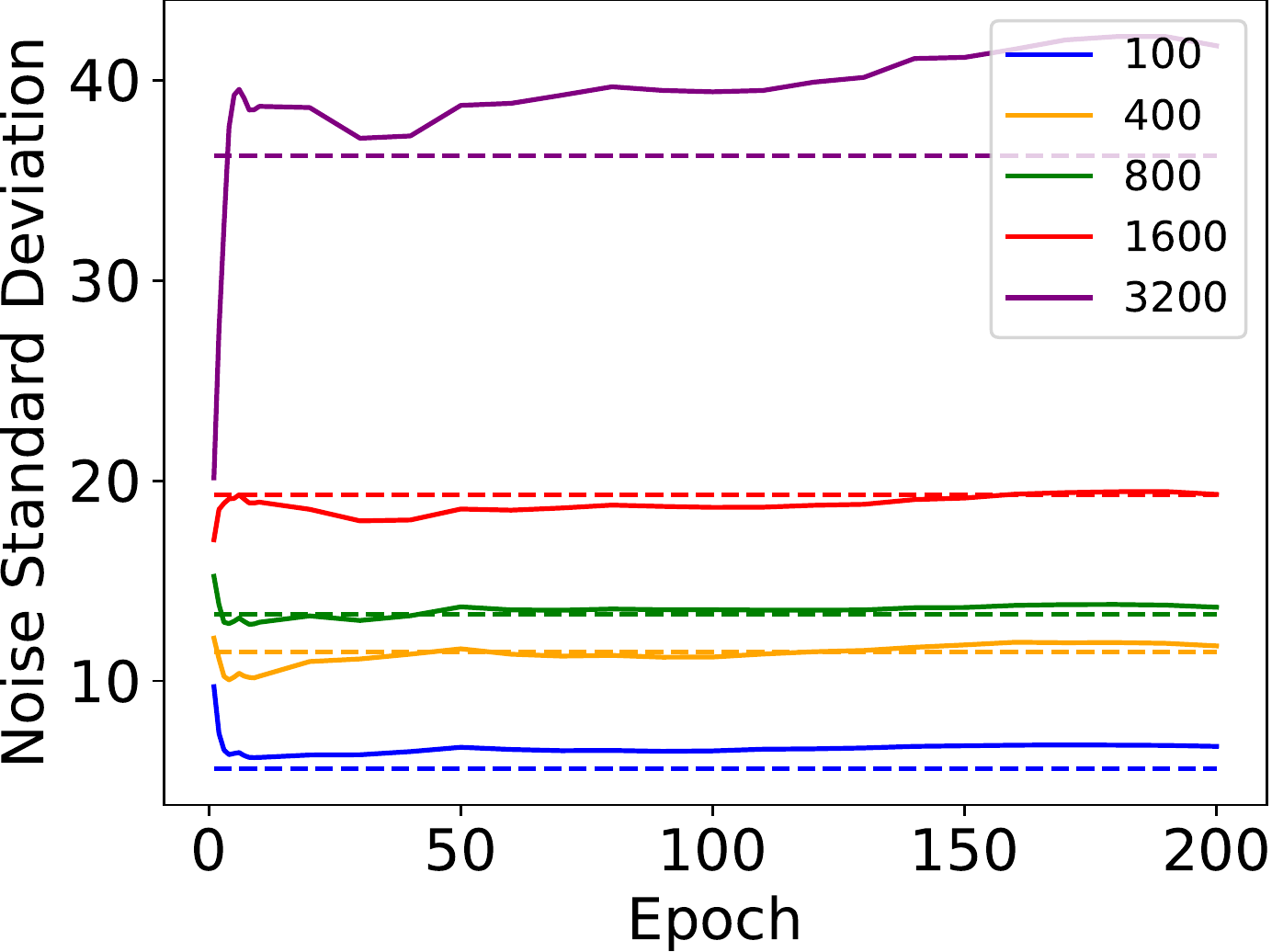} }}
    
    \caption{Standard deviation of learned noise model during training under (a) different cameras and (b) different ISO levels is close to the real ones. This shows the proposed conditional layers have learned to adjust the distribution based on those settings.  Dotted lines are the true values.}
    \label{fig:cam_iso}
\end{figure}

\paragraph*{Modeling Different Cameras and ISO Settings.}
Figure \ref{fig:cam_iso} shows the learning of noise characteristics under different cameras and ISO settings.
The dotted lines are the true noise standard deviation under each condition. The results show that the noise distribution changes drastically with different cameras and ISO levels.
Further, our model is able to successfully capture this behavior to learn a more realistic noise model.
The graphs also suggest that, while the model is quick to learn in the first few epochs and exhibits relatively little overfitting with the exception of ISO 3200.
However we note that this ISO setting has a very limited number of samples in the training set.

\begin{table}[t]
\centering
\begin{tabular}{ c c c }
    \hline
    Flow blocks & $NLL$ & $D_{KL}$ \\
    \hline
    CL & 3.678 & 0.079  \\
    CCS\textsubscript{ISO only}x2 & 3.726 & 0.154 \\
    CCS\textsubscript{camera only}x2 & 3.609 & 0.216 \\
    CCS\textsubscript{clean image only}x2 & 3.882 & 0.295 \\
    CCSx2 & 3.530 & 0.104 \\
    CL-CCSx2 & 3.398 & 0.075 \\
    (CL-CCSx2)\,x2 & 3.254 & 0.067 \\
    \hline
    (CL-CCSx2)\,x4 & \textbf{3.072} & \textbf{0.044} \\
     \hline
\end{tabular}
\caption{Test $NLL$ and $D_{KL}$ achieved by different flow steps. The symbols CL and CCS refer to the conditional linear and conditional coupling steps where a conditional coupling step is a combination of a 1x1 convolutional layer and a conditional affine coupling layer. The numbers next to x indicate the number of flow and coupling steps. Unless otherwise specified in the subscripts, the layers have the formulation mentioned in the methods sections. Last row is the architecture we use in other experiments.
}
\label{tab:ablation_studies}
\end{table}

% \begin{table}[h!]
% \centering
% \begin{tabular}{ c c c }
%     \hline
%     Model & $NLL$ & $D_{KL}$ \\
%     \hline
%     Lt2\_unconditional & 4.03 & 0.371 \\
%     Lt2\_only conditioned on iso & 3.718 & 0.119\\
%     Lt2\_only conditioned on cam & 3.831 & 0.201 \\
%     cond\_Lt2 & 3.67 & 0.079\\
%     Cond\_lt2-cac-cac & 3.397 & 0.082 \\
%     Cond\_lt2-cac-cac + cond\_lt2 & & \\
%     2 * (Cond\_lt2-cac-cac) + cond\_lt2 & 3.217 & 0.097 \\
%      \hline
%     4 * (Cond\_lt2-cac-cac) + cond\_lt2 & \textbf{3.036} & \textbf{0.055} \\
%      \hline
% \end{tabular}
% \caption{Conditional.}
% \label{tab:ablation_studies}
% \end{table}

\paragraph*{Ablation Studies.}
Table \ref{tab:ablation_studies} summarizes the performance of different architecture choices for the flow block of our normalizing flows model.
% Please note that the data processing step is the same for all proposed architectures in the table.
The results show a significant improvement in noise modeling and noise synthesis when the coupling step conditions on all the important variables including the clean image, camera type, and ISO settings, rather than conditioning on only one of them.
Additionally, we see an improvement by adding the conditional linear flow (CL) layer to the conditional coupling steps (CCS), showing the importance of having a direct way of transferring knowledge from camera types and ISO levels.
Finally, we show the importance of having multiple flow blocks. The architecture of (CL-CCSx2)\,x4 with four flow blocks achieves the best performance in both metrics, $NLL$ and $D_{KL}$.
This is the architecture used in other experiments.

\begin{table}[t]
\centering
% \begin{adjustbox}{width=\columnwidth,center}
\begin{tabular}{ c c c}
    \hline
    Noise Model & $PSNR$ & $SSIM$ \\
    \hline
     Isotropic Gaussian & 32.48 & 0.855 \\  
     Diagonal Gaussian & 33.34 & 0.867 \\ 
     Full Gaussian & 32.72 & 0.873 \\
     Heteroscedastic Gaussian & 32.24 & 0.849 \\
     Noise Flow & 33.81 & 0.894 \\
     C2N* &  33.76 & 0.901 \\
     Our model & \textbf{34.74} & \textbf{0.912} \\
     \hline 
    % \hdashline
     Real Noise & 36.51 & 0.922 \\
     \hline
\end{tabular}
% \end{adjustbox}
\caption{
Denoiser performance when trained on samples from each model. Denoisers are evaluated on the SIDD Benchmark set. The denoiser trained on samples from our noise model achieves better performance compared to those trained on noise from the baselines. 
(*) Results are taken from \cite{Jang2021C2n}.}
\label{tab:psnr_ssim_benchmark}
\end{table}

\subsection{Application: sRGB Denoising}
\label{application}
One of the main applications of noise modeling is to generate realistic noise to be used in downstream tasks, like denoising.
Here we explore the training of the standard DnCNN denoiser \cite{zhang2017beyond} using samples generated from the learned noise model to test its noise generation capabilities.

\paragraph*{Dataset}
To train DnCNN we use SIDD-Medium dataset.
Clean images are from SIDD-Medium and the noisy images are either the real ones provided with the dataset or generated by either the proposed noise model or one of the baseline models as specified.
% The baselines used in this section are the ones we introduced in the previous section.
We use noisy and clean images from SIDD-Validation and SIDD-Benchmark for validation and testing purposes, respectively. 

\paragraph*{Results}
Table \ref{tab:psnr_ssim_benchmark} summarizes the result of our denoising experiment.
The results show that a DnCNN model trained on the synthetic noises from our model achieves a significantly higher performance in terms of peak signal-to-noise ratio (PSNR) and structural similarity (SSIM) \cite{wang2004similarity} on SIDD-Benchmark compared to the denoisers trained on the samples from baseline models.
While the performance of our model does not exceed the performance of a denoiser trained with real data (\eg, as was found with noise models in RAW-rgb \cite{abdelhamed2019noiseflow}), it does significantly shrink the gap.
% Our model shrinks the gap between the denoisers trained on synthetic data and a denoiser trained on the real noisy image, indicated by DnCNN-Real in table \ref{tab:psnr_ssim_benchmark}.

\begin{figure}[ht]
    \centering
    \includegraphics[width=\columnwidth]{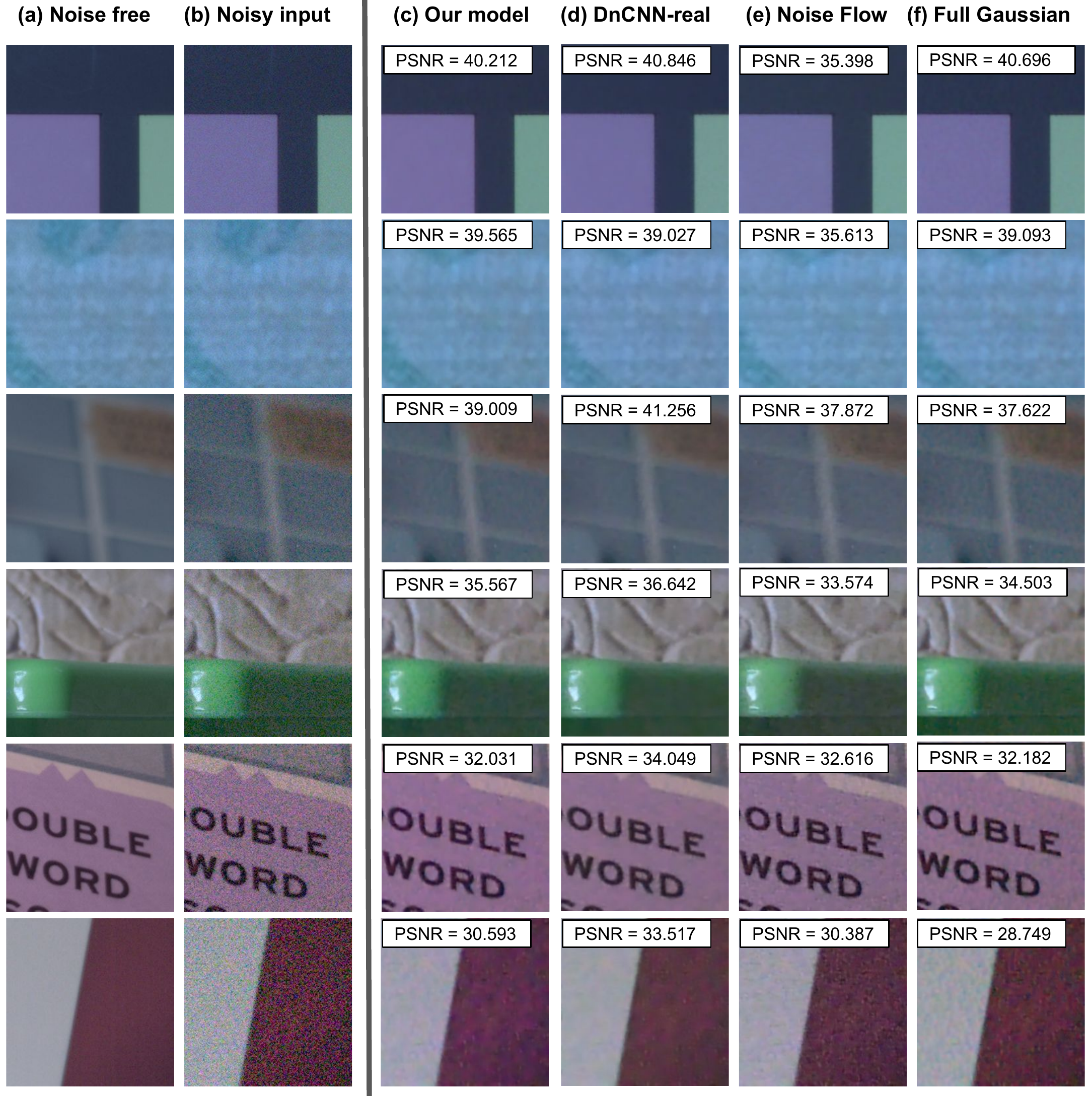}
    \caption{Denoising results on SIDD-Validation from denoisers trained on noisy images from (c) our model, (d) real noisy images of SIDD-Validation, and (e, f) two of our baselines. }
    \label{fig:denoising}
\end{figure}

Figure \ref{fig:denoising} shows denoising results of the DnCNN model trained with real noisy images and three noise synthesis strategies including our model and two of our baselines.
The figure also includes the input noisy image and the ground truth clean image for reference.
For the full set of results, we refer the readers to the supplemental materials.  The denoiser trained on noise samples from our model tends to produce denoised images that are closer to the ground truth clean images than when trained on samples from the baseline noise models.
The model trained on noisy image samples from Noise Flow tends to not remove the noise fully, outputting images which still contain a significant amount of noise (\eg, as in row 6).
This is likely caused by the tendency of Noise Flow to significantly overestimate the noise variance as demonstrated in Figure \ref{fig:red_var}.
Finally, denoisers trained on Gaussian samples tend to produce overly smooth denoised images, (\eg, as in row 4).
% Even though the DnCNN model trained on samples from our model does not get the highest PSNR in all cases, it is the most stable model across all six examples.

\section{Conclusion }
\label{sec:conclusion}
We introduced a noise model specifically tailored to capture image noise in the sRGB domain.
Because image noise in the sRGB domain exhibits significantly more complex noise distributions than those found in the unprocessed RAW-rgb domain, we showed that existing RAW-rgb noise models like heteroscedastic noise and the Noise Flow model \cite{abdelhamed2019noiseflow} are ineffective at capturing noise in sRGB.
To address this we described an architecture based on normalizing flows that can effectively model sRGB image noise, while capturing the complex dependencies on variables such as clean image intensity, camera model and ISO settings.
We demonstrated the effectiveness of the proposed noise model directly (with NLL and KL divergence metrics) and by training image denoisers using image noise synthesized by our model.
We showed that our image denoisers significantly outperform other denoisers trained on existing noise models for sRGB.
Source code is available at \url{https://yorkucvil.github.io/sRGBNoise/}.

\section*{Acknowledgments}
This work was done as part of an internship at the Samsung AI Center in Toronto, Canada.  AM's internship was funded by a Mitacs Accelerate.  SK's and AM's student funding came in part from the Canada First Research Excellence Fund for the Vision: Science to Applications (VISTA) programme and an NSERC Discovery Grant.  

%%%%%%%%% REFERENCES
{\small
\setlength{\bibsep}{0pt plus 0.3ex}
\bibliographystyle{IEEEtranSN}
\bibliography{references}
}

\newpage
\appendix
\section{Scale and Translation Functions}

Here we provide more details about the normalzing flow architecture used in the main paper.

\begin{figure}[ht]
    \centering
    \includegraphics[width=0.95\columnwidth]{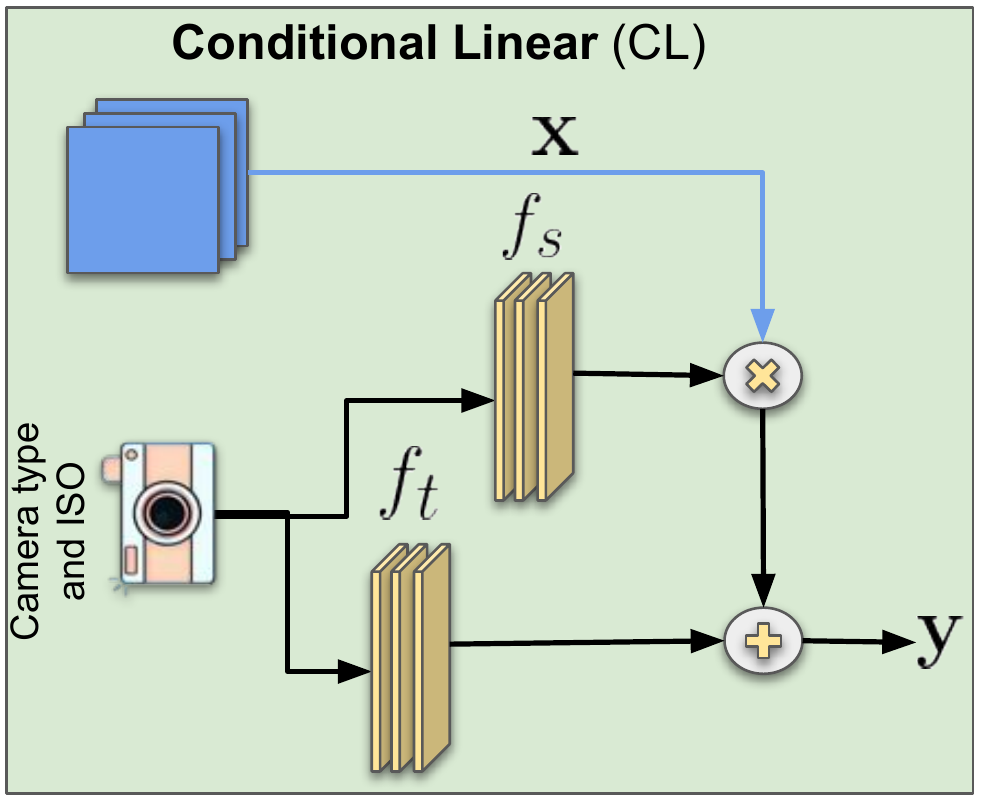}
    \caption{Forward pass of the conditional linear flow. Functions $f_{s}$ and $f_{t}$ are responsible for calculating the scale and bias terms, respectively. These terms are later applied to the input image $\dataval$ to calculate the output of this layer.} 
    \label{fig:ca_cg_arch}
\end{figure}
\paragraph*{Conditional Linear Flow} 
To condition on camera types and ISO levels, this layer creates a one-hot encoding of the camera-ISO pairs. For example, in our training and testing steps, we use images from five smartphones with five different ISO levels. This means the one-hot encoding of the pairs is a vector of size 25. $f_s: \mathbb{R}^{25} \xrightarrow{} \mathbb{R}^{3}$ and $f_t: \mathbb{R}^{25} \xrightarrow{} \mathbb{R}^{3}$ functions have one scale and bias parameter for each channel for a given camera-iso representation. Since we have 25 camera-ISO pairs and are working with three-channel sRGB data, each of the two functions has a set of parameters with shape $(25, 3)$. The two functions use the one-hot vectors to index into these parameters. As a result, the output of each function is a vector of size 3 (one value per channel). The values corresponding to each channel are later applied to all pixels in that channel.

\begin{figure}[t]
    \centering
    \includegraphics[width=0.95\columnwidth]{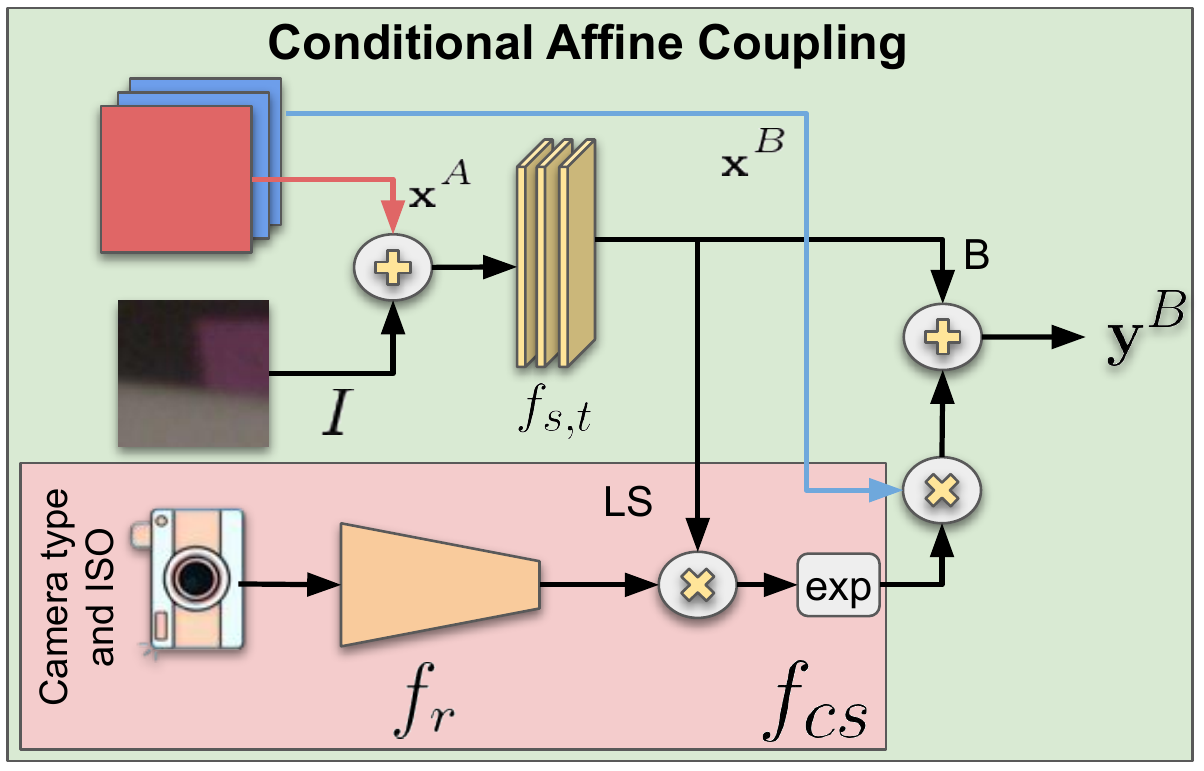}
    \caption{Forward pass of the conditional affine coupling layer. $f_{s,t}$ calculates the bias (B) and log\_scale (LS) from the clean image and one channel of the input data. A rescaling term is calculated from the camera type and ISO setting and applied to LS by $f_{cs}$ function. Later the rescaled LS is used along with the bias term and the remaining two channels of the input image to calculate $\transvalB$. $\transvalB$ is one part of the output.} 
    \label{fig:cac_arch}
\end{figure}
\paragraph*{Conditional Affine Coupling}

As shown in Figure \ref{fig:cac_arch}, to condition on all three variables, namely the clean image, camera type, and ISO level, $f_{cs}$ and $f_{ct}$ functions consider these variables in two steps. First, the clean image is concatenated with one subset of the input data and passed to a CNN model ($f_{s, t}$) to calculate the log scale and bias factors. Then, the camera type and gain setting are encoded into one-hot vectors separately. In our case, each encoding vector has a size of 5. The one-hot encoding vectors are concatenated and passed to a residual network ($f_{r}: \mathbb{R}^{10} \xrightarrow{} \mathbb{R}$) to calculate a rescaling factor. This rescaling factor is multiplied by the log scale factor calculated earlier to get the final scaling values. The bias term is returned untouched. These functions have the form:
\begin{align*}
    \mathbf{LS}, \mathbf{B} &= split(f_{s,t}(\datavalA, \cleanI)) \\
    f_{ct}(\datavalA|\cleanI,\camera,\ISO) &=  \mathbf{B} \\
    f_{cs}(\datavalA|\cleanI,\camera,\ISO) &= \exp(\mathbf{LS} * f_{r}(\camera, \ISO))
\end{align*}
where $f_{e}$ is the one-hot encoder, and $\mathbf{LS}$ and $\mathbf{B}$ are the log scale and bias factors, respectively. 

\section{Inverse Gamma}
RAW-rgb images go through an in-camera imaging pipeline that transforms the image from the RAW-rgb space to the sRGB domain. The steps in this pipeline introduce non-linearities that result in a complex noise distribution in the sRGB space. One of the main nonlinear steps in this pipeline is gamma correction which is an invertible process. Its inverse is commonly used to approximately linearize sRGB data and can easily be implemented as a normalizing flow transformation. Inverse gamma is defined as $\transval = \dataval^{\gammaval}$ where the default value of $\gammaval$ is 2.2. In this section, we explore the effects this layer has on Noise Flow and our proposed model when added to the data processing step.

\begin{figure}[ht]
    \centering
    \includegraphics[width=0.95\columnwidth]{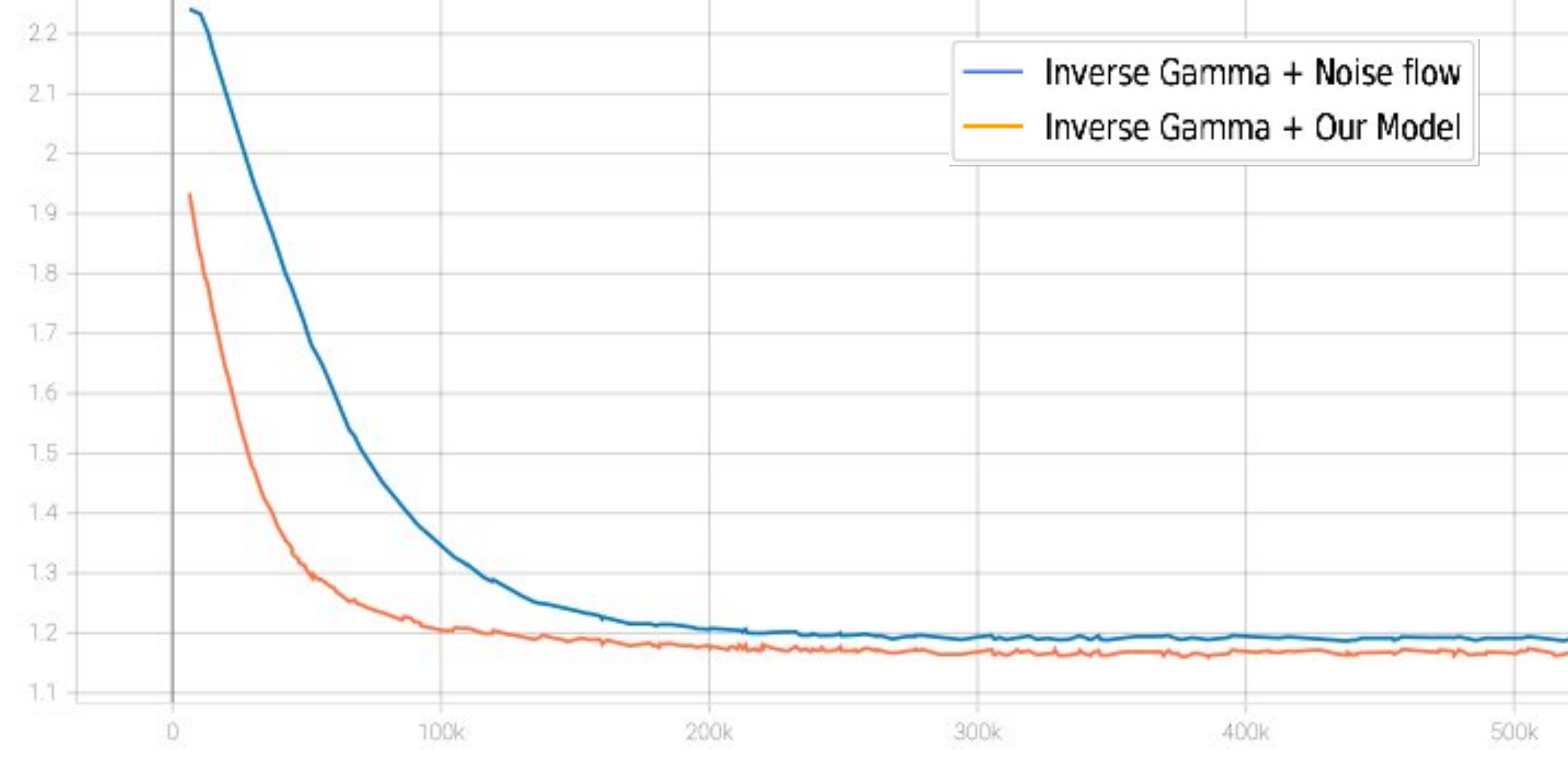}
    \caption{$\gammaval$ value changes in the training process. $\gammaval$ converges to 1.2 for both models.}
    \label{fig:inverse_gamma_train}
\end{figure}
\begin{table}[ht]
\centering
\begin{adjustbox}{width=\columnwidth,center}
\begin{tabular}{ c c c c }
    \hline
    Model & $NLL$ & $D_{KL}$ & $\# Params$ \\
    \hline
     Noise flow & 3.311 & 0.198 & 2330\\
     Our model & \textbf{3.072} & \textbf{0.044} & 6160 \\
     \hline
    Inverse gamma + Noise flow & 3.322 & 0.181 & 2332 \\
    Inverse gamma + Our model & 3.092 & 0.061 & 6162 \\
     \hline
\end{tabular}
\end{adjustbox}
\caption{Models with the inverse gamma transformation added to their data processing step have similar performances as the original models.}
\label{tab:inverse_gamma}
\end{table}

Table \ref{tab:inverse_gamma} shows that using the inverse gamma layer does not improve the modeling and sampling performances.  Figure \ref{fig:inverse_gamma_train} shows that $\gammaval$ gamma converges to 1.2 from the initial value of 2.2 resulting in a near identity transformation for both models.

\section{Architecture Search}
There are many more approaches and flow based transformations that can be used to form the flow block of our model. Here we introduce a few more novel conditional transformations and explore their modeling capabilities.

\subsection{Conditional Transformations}

\begin{figure}[ht]
    \centering
    \includegraphics[width=0.95\columnwidth]{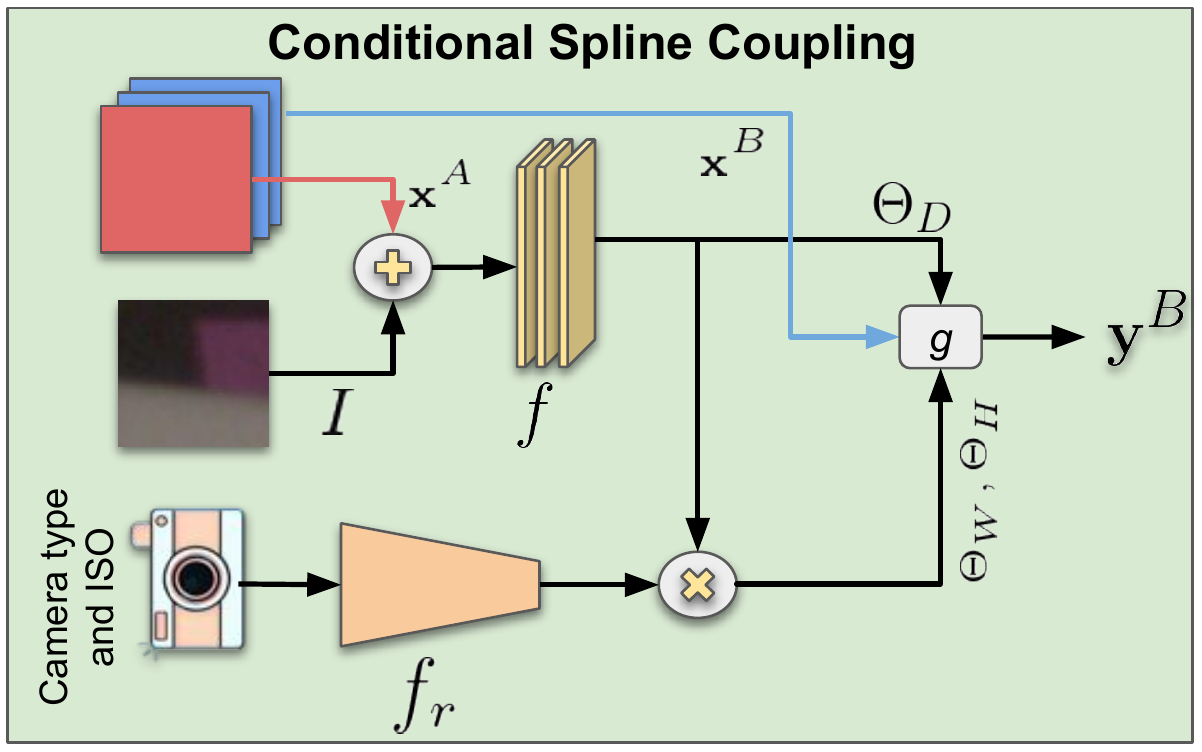}
    \caption{Forward pass of the conditional spline coupling layer.
    Function $f$ calculates the bins' parameters from the clean image and one of the input channels. Later some of these parameters are rescaled using the rescale term calculated by $f_r$ from the one-hot encodings of camera and gain settings. Finally, the updated parameters are passed to function $g$ introduced in the neural spline paper \cite{durkan2019neural}. This function is responsible for calculating the output of this layer.} 
    \label{fig:csc_arch}
\end{figure}
\paragraph*{Conditional Spline Coupling (CSC)}
This layer is an extension of the neural spline layer~\cite{durkan2019neural}. We extend the neural spline transformation to condition on the clean image, camera type and gain setting. Here, we outline the procedure to condition on these variables:

As shown in Figure \ref{fig:csc_arch}, the clean image, $I$, is concatenated channel-wise with $\datavalA$ and passed to function $f$ to calculate the parameters of the bins. Then, similar to the conditional affine coupling layer, the camera and gain settings are encoded into one-hot vectors separately. $f_r: \mathbb{R}^{10} \xrightarrow{} \mathbb{R}$, which is an arbitrary function, takes the resulting vectors as input and outputs a scale factor. In our experiments we use a residual network architecture. The scale factor is later used to rescale the bins' width and height parameters.
The conditional spline coupling layer is defined as follows:
\begin{align*}
    \Theta_W, \Theta_H, \Theta_D = f(\datavalA, I) \\
    \Theta_W, \Theta_H = (\Theta_W, \Theta_H) * f_{r}(\camera, \ISO) \\
    \transvalB = g(\datavalB|\Theta_W, \Theta_H, \Theta_D)
\end{align*}
where function $g$ is the same function used in the standard neural spline flows. The inverse and log determinant of this layer can be calculated by following the same procedure used by the unconditional neural spline layer.

Similar to the affine coupling layer, the step that conditions on the clean image is independent of the step that conditions on the camera and gain. This allows us to use this layer in multiple ways in our experiments in Section \ref{experiments}.

\paragraph*{Conditional Affine (CA)}
The nature of the noise and the subsequent non-linear processing is heavily determined by the underlying noise free image and the specific camera and gain (or ISO) settings used.
To account for this, we introduce linear flow layers, which are conditioned on combinations of important variables including the camera, $\camera$, gain setting, $\ISO$, and the underlying clean image, $\cleanI$. This layer comes in two forms depending on which subset of variables it is considering:

\begin{figure}[t]
    \centering
    \includegraphics[width=0.95\columnwidth]{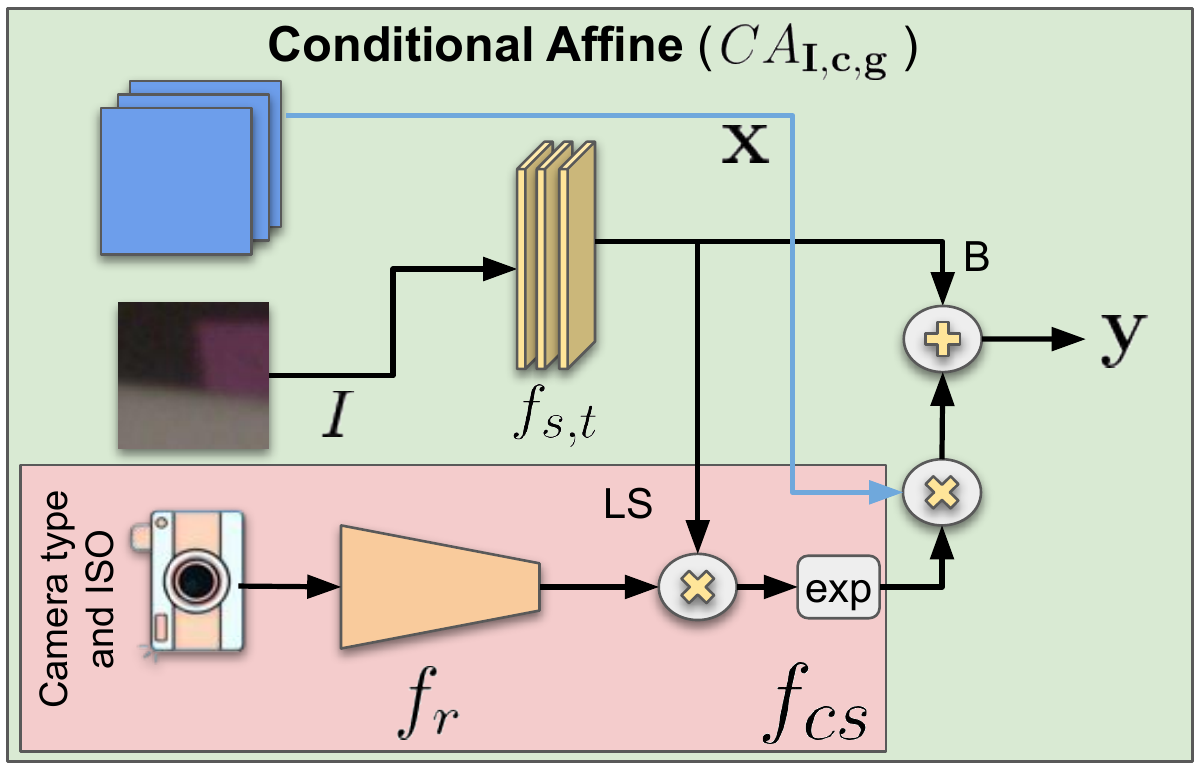}
    \caption{Forward pass of the conditional affine layer conditioned on the clean image, camera type and gain signal. $f_{s,t}$ calculates the bias (B) and log\_scale (LS) from the clean image. The camera type and ISO setting are used to learn a rescaling factor which is later applied to LS by $f_{cs}$ function. Finally,  the rescaled LS is used along with the bias term and the input image to calculate $\transval$.} 
    \label{fig:ca_icg_arch}
\end{figure}
\textbf{Conditioned on Clean Image, Camera and ISO ($CA_{\cleanI, \camera, \ISO}$):} 
This layer is similar to the conditional affine coupling layer. The only difference is that this layer does not split the input dimension. It calculates the scale and bias and applies them to all input dimensions. This layer has the following form:
\begin{equation*}
\transval = \dataval \odot f_{cs}(\cleanI,\camera,\ISO) 
    + f_{ct}(\cleanI,\camera,\ISO),
\end{equation*}
where $\dataval$ and $\transval$ are the input and output of this layer, respectively. $f_{cs}$ and $f_{ct}$ are similar to the functions used by the affine coupling layer with not taking $\datavalA$ as an input being the only difference. As a result, $f_{s, t}$, which is defined earlier, only uses the clean image to calculated the log scale and bias terms calculated. Later a rescaling factor is generated from the encoding of the camera type and ISO setting and applied to the log scale term.
Similar to the CAC layer, the inverse and log determinant of this transformation can be easily calculated.

\begin{figure}[t]
    \centering
    \includegraphics[width=0.95\columnwidth]{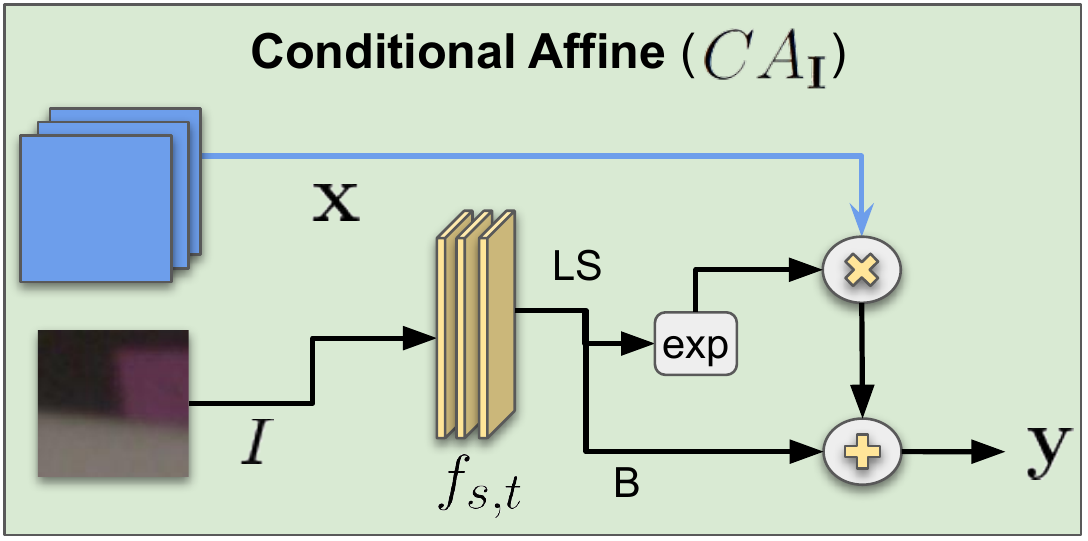}
    \caption{Forward pass of the conditional affine layer conditioned on the clean image. $f_{s,t}$ calculates the bias (B) and log\_scale (LS) from the clean image. These factors are applied directly to the input image to calculate the output of this layer.} 
    \label{fig:ca_i_arch}
\end{figure}

\textbf{Conditioned on Clean Image ($CA_{\cleanI}$):}
$f_{cs}$ and $f_{ct}$ of $CA_{\cleanI, \camera, \ISO}$ use a two step process to calculate the scale and bias terms. First, they calculated the log scale and bias terms. Then, a rescale term is learned from variables such as gain and applied to the scale factor.  These two steps are independent of each other and can be applied separately. In this layer, we only use the first step to learn the log scale and bias terms from the clean image. This layer has the following form:
\begin{align*}
    \mathbf{LS}, \mathbf{B} &= split(f_{s,t}((\datavalA, \cleanI)) \\
    \transval &= \dataval \odot \exp{(\mathbf{LS})} 
    + \mathbf{B}
\end{align*}

Since the second step of $f_{cs}$ and $f_{ct}$ is not applied in this layer, information learned from the clean image is directly used to scale and shift the input data with out being rescaled. In Section \ref{experiments} we investigate whether such direct transfer of information from the underlying clean representation to the noise image is beneficial.

\subsection{Experiments}
\label{experiments}
In this section, we consider multiple architectures for the flow block . The goal is to understand what combination of layers results in better modeling and generative performance. In our experiments, unless otherwise specified, all coupling layers are preceded by an invertible 1x1 convolution layer. Additionally, the coupling layers come in pairs. As a result, we have "x2" next to the coupling layers in our tables.

First, we investigate the effects of unconditional layers. Table \ref{tab:unconditional_experiment} summarizes the results of experiment. The model defined in the first row uses two spline coupling layers as the unconditional step followed by some conditional layers. The model described in row two is similar to the first model with the exception that unconditional layers are eliminated. Even though this model has fewer parameters due to the lack of unconditional layers, it outperforms the first model in both metrics. Eliminating the unconditional layers results in a 1.162 nats/pixel improvement of NLL.
The third row replaces the unconditional/conditional spline coupling transformations with unconditional/conditional affine coupling layers. A similar pattern of improvement emerges when this model is compared with the last row where the unconditional layers are removed. Our last model achieves an NLL of 3.602 which translates to a 0.545 nats/pixel improvement.
Additionally, when compared to the second row, the last row reveals that the conditional affine coupling (CAC) layer is a better fit in terms of $D_{KL}$ than the conditional spline couplin (CSC) layer for conditioning on the clean image, camera type, and gain setting.
As a result, we use the model in row four as a base for our next experiment.

\begin{table}[t]
\centering
\begin{adjustbox}{width=\columnwidth,center}
\begin{tabular}{ c c c c c c c }
    \hline
    Uncond. & $\ISO$ and $\camera$ & $\cleanI$ & $\cleanI$, $\ISO$, and $\camera$ & $NLL$ & $D_{KL}$  \\
    \hline
    SCx2 & CL & $\textrm{CA}_{\cleanI}$ & CSCx2 & 4.238 & 0.229 \\ % 1_nsx2_cl_ca_cnsx2
    - & CL & $\textrm{CA}_{\cleanI}$ & CSCx2 & \textbf{3.076} & 0.141 \\ % 1_-_cl_ca_cnsx2
    ACx2 & CL & $\textrm{CA}_{\cleanI}$ & CACx2 & 4.147 & 0.183 \\ % 1_acx2_cl_ca_cacx2
    - & CL & $\textrm{CA}_{\cleanI}$ & CACx2 & 3.602 & \textbf{0.104} \\ %  1_-_cl_ca_cacx2
    \hline
\end{tabular}
\end{adjustbox}
\caption{\textbf{Unconditional layer experiment.} Models in this table use 1 flow block (S = 1). Removing the unconditional layer and keeping conditional layers untouched improves the performance of the models. This trend is independent of the type of unconditional layer used in these models. 
Models that use unconditional/conditional affine coupling layers achieve better $D_{KL}$ compared to the models that use unconditional/conditional spline coupling layers. A lower $D_{KL}$ suggests better generated samples.}
\label{tab:unconditional_experiment}
\end{table}

The next experiment is designed to show the effects of layers only conditioned on the clean image. The three models shown in Table \ref{tab:only_clean_experiment} are identical with their choice of layer for conditioning on the clean image being the only difference. The results suggest removing the layer that only conditions on the clean image improves the performance in terms of $D_{KL}$ but might worsen the $NLL$ results. We find $D_{KL}$ to be a better indicator of the quality of the samples and prefer simpler model. Therefore, we conclude that there is no need to have a specialized layer to solely condition on the clean image.
As shown in Table \ref{tab:only_clean_experiment} the model defined in row three achieves a $D_{KL}$ value of 0.077 which is the best result achieved so far. As a result, we include this model in Table \ref{tab:iso_cam_experiment} to help with our next experiment.

\begin{table}[t]
\centering
% \begin{adjustbox}{width=\columnwidth,center}
\begin{tabular}{ c c c c c }
    \hline
    $\ISO$ and $\camera$ & $\cleanI$ & $\cleanI$, $\ISO$, and $\camera$ & $NLL$ & $D_{KL}$ \\
    \hline
    CL & $\textrm{CA}_{\cleanI}$ & CACx2 & 3.602 & 0.104 \\ %  1_-_cl_ca_cacx2
     CL & CSCx2 & CACx2 & \textbf{2.901} & 0.163 \\ % 1_-_cl_cnsx2_cacx2
    CL & - & CACx2 & 3.473 & \textbf{0.077} \\ % 1_-_cl_-_cacx2
    \hline
\end{tabular}
% \end{adjustbox}
\caption{\textbf{Clean image only experiment.}
Models in this table use 1 flow block (S = 1).
Adding a specialized layer to only condition on the clean image harms $D_{KL}$ but it might improve (reduce) NLL. We consider two specialized layers for conditioning on the clean image, namely $\textrm{CA}_{\cleanI}$ and CSCx2. They both fail to improve the model performance in terms of $D_{KL}$.}
\label{tab:only_clean_experiment}
\end{table}

\begin{table}[t]
\centering
\begin{adjustbox}{width=\columnwidth,center}
\begin{tabular}{ c c c c }
    \hline
    ISO and Cam & Clean, ISO, and Cam & $NLL$ & $D_{KL}$ \\
    \hline
    CL & CACx2 & 3.473 & 0.077 \\ % 1_-_cl_-_cacx2
    CL & CSCx2 & \textbf{3.064} &0.159 \\ % 1_-_cl_-_cnsx2
    CL & $\textrm{CA}_{\cleanI, \camera, \ISO}$ & 3.636 & \textbf{0.063} \\ % 1_-_cl_-_ca
    CACx2 & $\textrm{CA}_{\cleanI, \camera, \ISO}$ & 3.522 & 0.099 \\ % 1_-_cacx2_-_ca
    \hline
\end{tabular}
\end{adjustbox}
\caption{\textbf{Conditioning on all variables.}
Models in this table use 1 flow block (S = 1).
The choice of transformation for conditioning on ISO, camera type, and clean image has a significant impact on the overall performance of the normalizing flows model. CL seems to be the best choice for conditioning on gain and camera as the models containing this layer achieve the best $NLL$ and $D_{KL}$.}
\label{tab:iso_cam_experiment}
\end{table}

The previous experiments show that having an unconditional layer and a layer that only conditions on the clean image is not necessary. The third experiment focuses on the conditional transformations that either utilize ISO and camera type information or use all three variables. 
Table \ref{tab:iso_cam_experiment} summarizes the experiment and shows CL\_CACx2 model and CL\_$\textrm{CA}_{\cleanI, \camera, \ISO}$ with one flow block achieve the best $D_{KL}$ performances with values of 0.077 and 0.063, respectively. Given the architectural design choices we made, the models can easily be made deeper by increasing the number blocks.

Table \ref{tab:final_models_experiment} shows the performance of these two models when the number of blocks is 4. The results suggest making a model deeper improves the results. For example, CL\_CACx2 with 4 repeated flow blocks achieves the best NLL with 0.401 nats/pixel improvement over the same model architecture with only one flow block.
CL\_CACx2 model with four blocks outperforms the other models in this table in both metrics. It achieves $NLL$ and $D_{KL}$ of 3.072 and 0.044, respectively.
This is the same model as the model we introduced in the paper.
% Given that this model achieves the best $D_{KL}$ among the models we investigate, we focus on this model for the remaining experiments in the following sections. 

\begin{table}[t]
\centering
\begin{adjustbox}{width=\columnwidth,center}
\begin{tabular}{ c c c c c }
    \hline
    \# Flow Blocks (S) & $\ISO$ and $\camera$ & $\cleanI$, $\ISO$, and $\camera$ & $NLL$ & $D_{KL}$ \\
    \hline
    1 & CL & CACx2 & 3.473 & 0.077 \\ % 1_-_cl_-_cacx2
    1 & CL & $\textrm{CA}_{\cleanI, \camera, \ISO}$ & 3.636 & 0.063\\ % 1_-_cl_-_ca
    \hline
    4 & CL & CACx2 & \textbf{3.072} & \textbf{0.044} \\
    4 & CL & $\textrm{CA}_{\cleanI, \camera, \ISO}$ & 3.639 & 0.060\\ % 4_-_cl_-_ca
    \hline
\end{tabular}
\end{adjustbox}
\caption{\textbf{Model depth.} Increasing the number of flow blocks improves the noise modeling capabilities of the NF models. The model in row three achieves the best performance in terms of $D_{KL}$ compared to other models.}
\label{tab:final_models_experiment}
\end{table}

\begin{figure*}[t]
    \centering
    \includegraphics[width=0.99\textwidth]{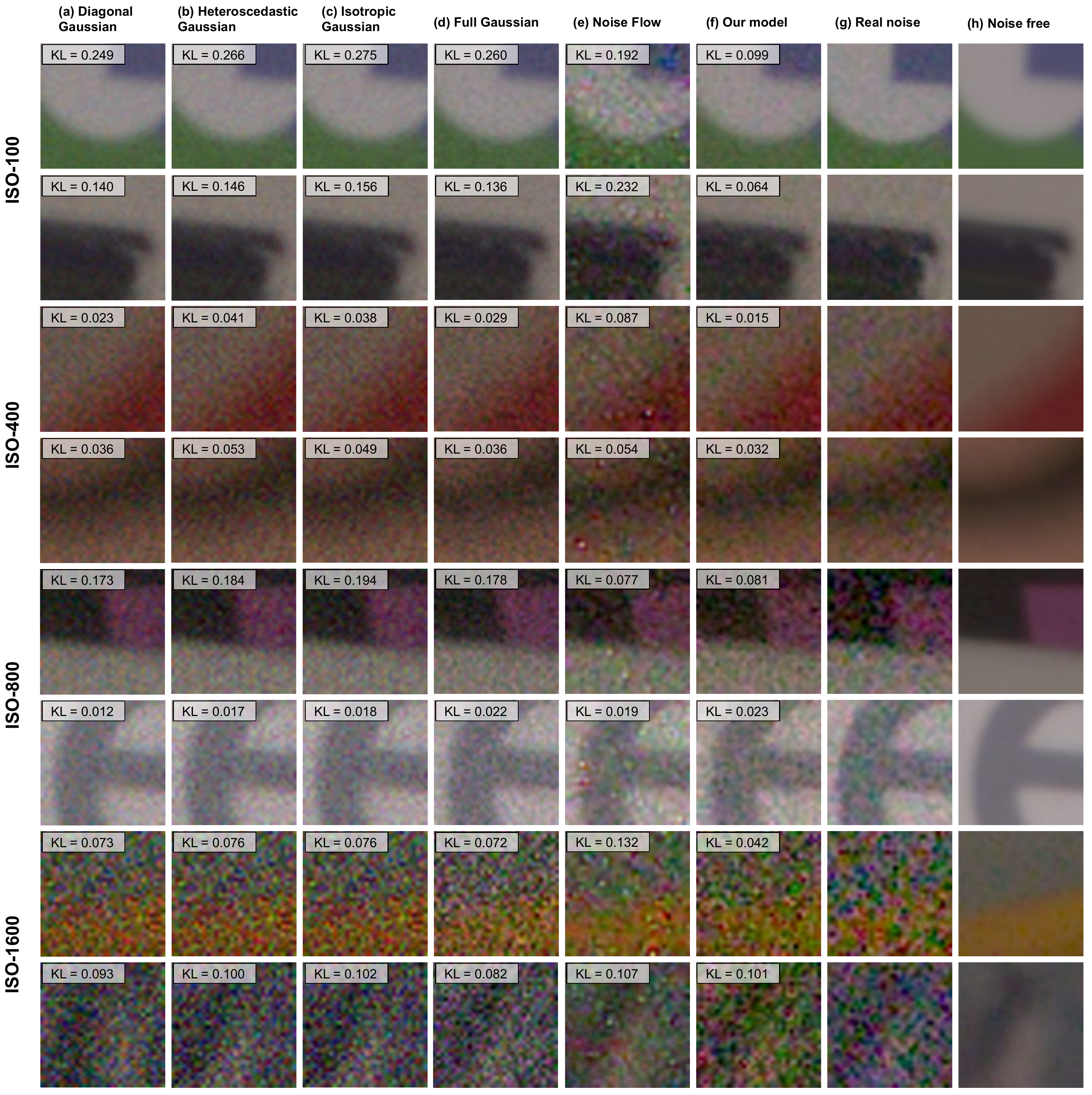}
    \caption{Generated samples from our model and all baselines. Samples from our model have noticeable visual similarities with the real noisy samples. Our samples achieve the lowest $D_{KL}$ in almost all cases, showing its ability to generate realistic noise.}
    \label{fig:noisy_images_all}
\end{figure*}
\begin{figure*}[ht]
    \centering
    \includegraphics[width=0.99\textwidth]{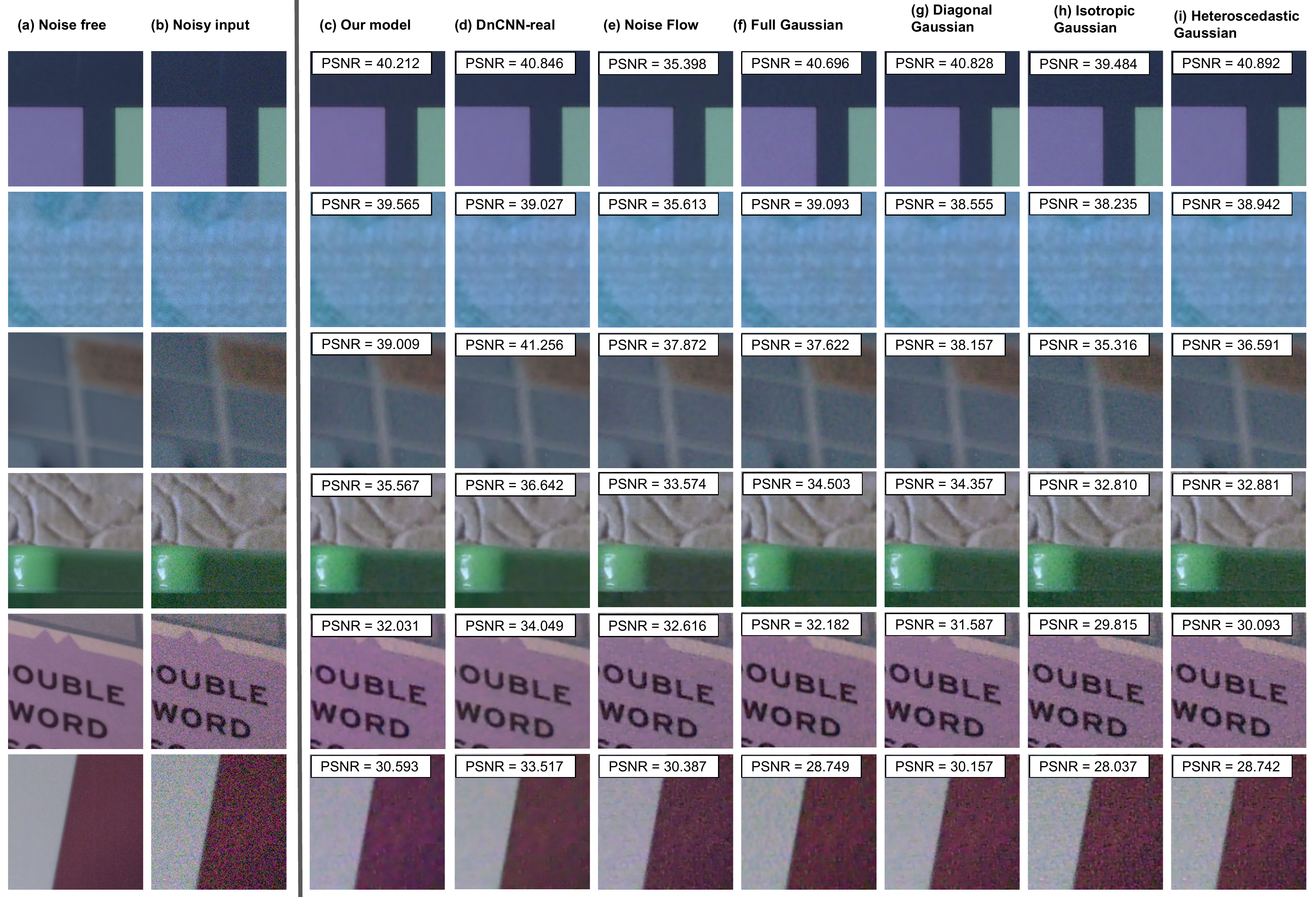}
    \caption{Denoising results on SIDD-Validation from denoisers trained on noisy images from (c) our model, (d) real noisy images of SIDD-Validation, and (e, i) all of our baselines. }
    \label{fig:denoising_all}
\end{figure*}

\end{document}